\documentclass[runningheads]{llncs}

\usepackage{eccv}

\usepackage{eccvabbrv}
\usepackage{graphicx}
\usepackage{booktabs}
\usepackage{algorithm}
\usepackage{xcolor}
\usepackage{colortbl}
\usepackage{multirow}
\usepackage{amsmath}
\usepackage{makecell}
\usepackage{bbm}
\usepackage{tabularx}
\usepackage[accsupp]{axessibility}  
\definecolor{eccvblue}{rgb}{0.21,0.49,0.74}
\definecolor{red}{rgb}{0.90,0.0,0.20}
\definecolor{green}{rgb}{0.49,0.76,0.00}
\definecolor{rulecolor}{HTML}{888888}
\definecolor{first}{HTML}{a9c98f}
\definecolor{second}{HTML}{cadeba}
\definecolor{third}{HTML}{eaf2e3}
\definecolor{beige}{HTML}{f2f0e9}
\definecolor{sunny}{HTML}{fffce8}
\newcommand{\method}{Geo-ID\xspace}
 
\newcommand{\cI}{\cellcolor{first}}
\newcommand{\cII}{\cellcolor{second}}
\newcommand{\cIII}{\cellcolor{third}}
\newcommand{\cg}{\cellcolor{lightgray}}

\newcommand{\impr}{\cellcolor{sunny}}
\usepackage{amsmath}
\usepackage{colortbl}
\usepackage{makecell}

\arrayrulecolor{rulecolor}
\DeclareUnicodeCharacter{2194}{\ensuremath{\leftrightarrow}}
\usepackage{listings}

\usepackage[pagebackref,breaklinks,colorlinks,citecolor=eccvblue]{hyperref}

\usepackage{orcidlink}
\newcommand{\rgbx}{RGB$\leftrightarrow$X}

\begin{document}

\title{Geo-ID: Test-Time Geometric Consensus for Cross-View Consistent Intrinsics} 

\titlerunning{Geo-ID}

\author{Alara Dirik\inst{1}\orcidlink{0000-0002-4946-1313}  \and
Stefanos Zafeiriou\inst{1}\orcidlink{0000-0002-5222-1740}}

\authorrunning{Dirik et al.}

\vspace{-.5em}
\institute{Imperial College London, UK \\
\email{\{a.dirik22,s.zafeiriou\}@imperial.ac.uk} \\
Project page: \href{https://alaradirik.github.io/geoid/}{https://alaradirik.github.io/geoid/}
}

\maketitle

\vspace{-1em}
\begin{center}
  \includegraphics[width=.9\linewidth]{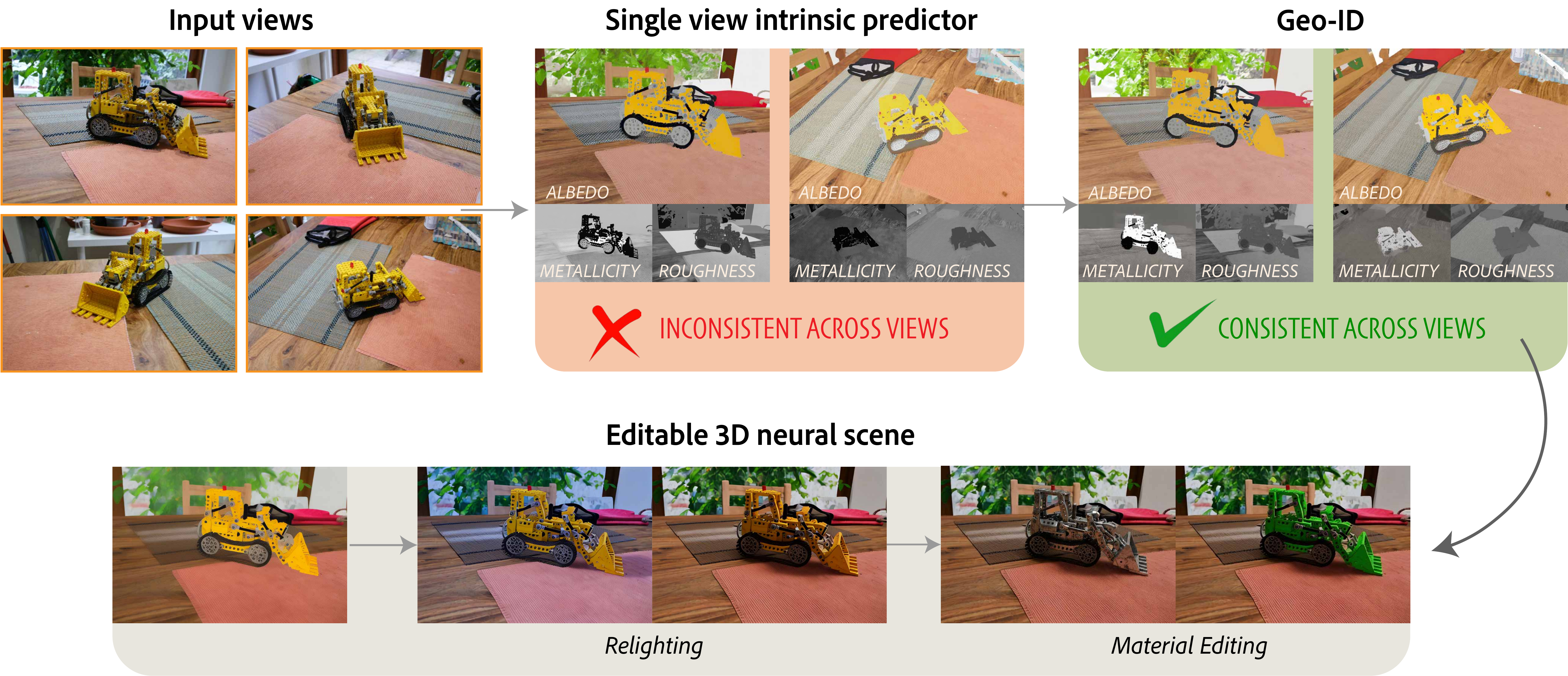}
\captionof{figure}{Existing single-view intrinsic predictors estimate PBR parameters independently per view, leading to cross-view inconsistencies. \method produces consistent decompositions by coupling per-view predictions through geometric correspondences at test time, enabling coherent material editing and relighting in downstream neural scene representations.}
  \label{fig:teaser}
\end{center}
\vspace{-1.5em}

\begin{abstract}
Intrinsic image decomposition aims to estimate physically based rendering (PBR) parameters such as albedo, roughness, and metallicity from images. While recent methods achieve strong single-view predictions, applying them independently to multiple views of the same scene often yields inconsistent estimates, limiting their use in downstream applications such as editable neural scenes and 3D reconstruction. Video-based models can improve cross-frame consistency but require dense, ordered sequences and substantial compute, limiting their applicability to sparse, unordered image collections. We propose Geo-ID, a novel test-time framework that repurposes pretrained single-view intrinsic predictors to produce cross-view consistent decompositions by coupling independent per-view predictions through sparse geometric correspondences that form uncertainty-aware consensus targets. Geo-ID is model-agnostic, requires no retraining or inverse rendering, and applies directly to off-the-shelf intrinsic predictors. Experiments on synthetic benchmarks and real-world scenes demonstrate substantial improvements in cross-view intrinsic consistency as the number of views increases, while maintaining comparable single-view decomposition performance. We further show that the resulting consistent intrinsics enable coherent appearance editing and relighting in downstream neural scene representations. 
\end{abstract}



\section{Introduction}
\begin{figure*}
    \centering
    \includegraphics[width=\textwidth]{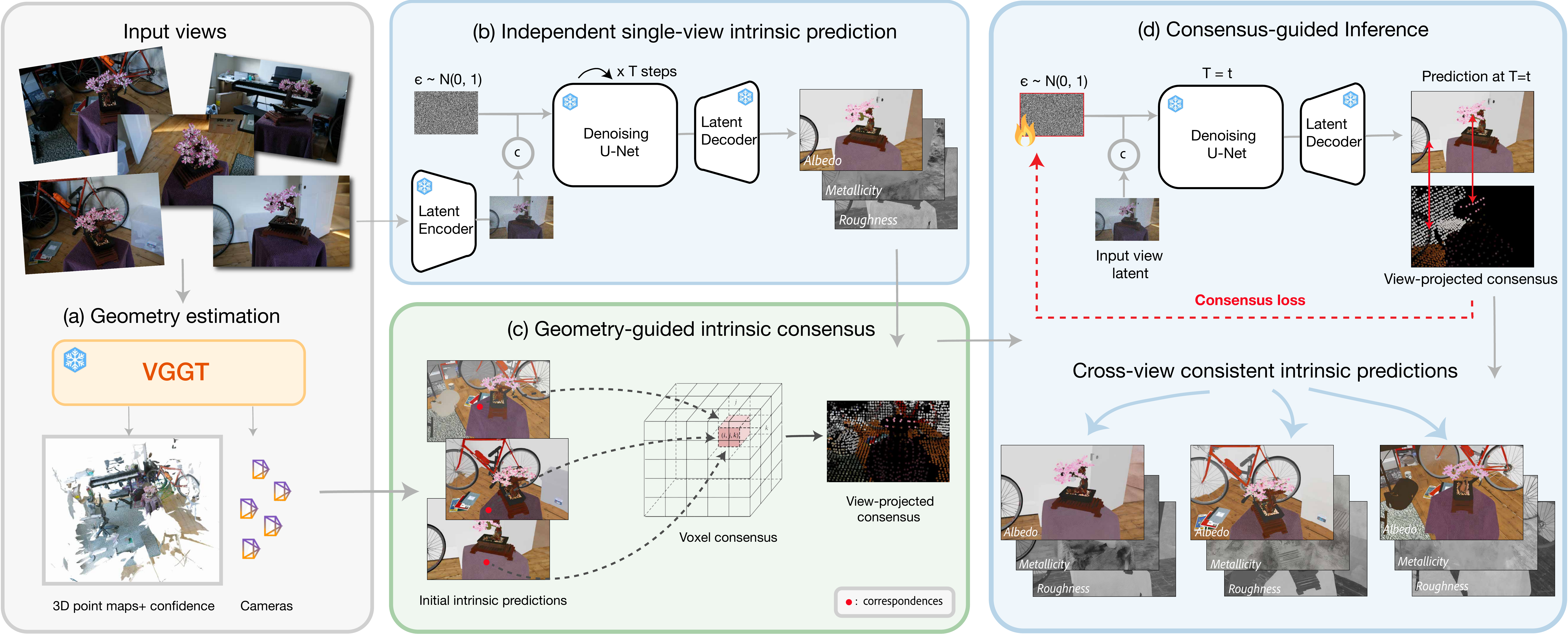}
    \caption{%
    \textbf{Overview of \method.}
    The pipeline consists of three phases. (1) Geometry-guided correspondence estimation: (a) We use a pretrained geometry transformer (VGGT) to predict camera parameters and dense 3D point maps with confidence from the input views. (2) Voxel-based intrinsic consensus: (b) A pretrained single-view diffusion model first produces independent intrinsic predictions for each view. (c) We voxelise high-confidence 3D points and aggregate the corresponding intrinsic values into a robust voxel-level consensus, which we reproject into each image. (3) Consensus-guided diffusion: (d) We run a second diffusion pass per view and inject the view-projected consensus as sparse guidance at selected denoising steps, producing cross-view consistent intrinsic predictions.
    }
    \label{fig:pipeline}
    \vspace{-1em}
\end{figure*}

Intrinsic image decomposition is a long-standing computer vision problem that aims to recover illumination-invariant surface properties such as albedo, roughness, and metallicity from images. Recent diffusion-based and vision transformer models~\cite{Careaga2024ColorfulDI,Dirik2025PRISMAU,Fu2024GeoWizardUT,Zeng2024RGBXID,Ke2025MarigoldAA,Kocsis2024IntrinsicID} achieve high-quality predictions from a single view and generalize well across diverse scenes.
However, when applied independently to multiple views of the same scene, these models often produce highly inconsistent intrinsic estimates. This inconsistency arises from a fundamental ambiguity: intrinsic decomposition from a single image is severely ill-posed, as geometry, illumination, and material appearance are entangled. Without explicit geometric context, multiple decompositions may be equally plausible, causing the same surface to be assigned different albedo or roughness values depending on viewpoint. Such inconsistencies are a major obstacle to building editable neural scene representations from sparse image collections, where coherent material predictions across views are essential.

Prior work has addressed this issue through two main directions. Video-based intrinsic decomposition models exploit temporal coherence to regularize predictions across frames~\cite{Liang2025DiffusionRendererNI,Sun2025OuroborosSD}, while multi-view methods either employ architectures trained jointly on multi-view data~\cite{Du2025IDTAP} or rely on inverse rendering pipelines that optimize geometry and materials per scene~\cite{Zhang2021NeRFactor,Liang2023GSIR3G,Kocsis2025IntrinsicIF}. While effective for dense captures or controlled settings, these approaches require specialized training data, heavy per-scene optimization, or accurate and complete geometry, assumptions that rarely hold for sparse, unordered image collections commonly encountered in practice.

In this work, we take a different perspective. Rather than designing a new intrinsic predictor or performing per-scene optimization, we introduce \method, an inference-time framework that repurposes pretrained single-view intrinsic models for cross-view consistent decomposition. Our key insight is that sparse geometric correspondences provide sufficient structure to resolve cross-view ambiguities, even when geometry is noisy or incomplete. By coupling independent per-view predictions through geometry-induced consensus constraints, we enforce agreement on shared surface properties without requiring dense geometry, retraining, or joint multi-image optimization, while preserving the original generative prior of the base model. 

\method operates entirely at test time and is applicable to existing diffusion-based intrinsic predictors. We demonstrate the approach using RGB$\leftrightarrow$X~\cite{Zeng2024RGBXID} and Marigold IID Appearance~\cite{Ke2025MarigoldAA}, and evaluate on synthetic benchmarks with ground-truth PBR parameters as well as real-world multi-view scenes. Our results show that \method substantially improves cross-view intrinsic consistency as the number of views increases, while maintaining comparable single-view decomposition quality. In summary, our contributions are:

\begin{itemize}
    \item \method, an inference-time framework that produces cross-view consistent intrinsic decompositions from sparse, unordered image collections by coupling frozen single-view diffusion models through geometric correspondences, without retraining or per-scene optimization.
    \item A geometry-driven consensus formulation that aggregates per-view intrinsic predictions through sparse 3D correspondences using robust, confidence-weighted statistics.
    \item A consensus-guided diffusion strategy that injects sparse cross-view constraints into the denoising process while preserving the generative prior of the base model.
\end{itemize}

\section{Related Work}

\subsection{Intrinsic Image Decomposition}

Intrinsic image decomposition seeks to separate an image into geometry-, reflectance-, and illumination-related components, and more recently into physically motivated PBR parameters such as albedo, roughness, and metallicity.
Early approaches relied on Retinex-style assumptions or optimization over hand-crafted priors~\cite{Land1971LightnessAR,Barron2012ShapeAA,Gehler2011RecoveringII}. Subsequent learning-based methods introduced convolutional and transformer architectures trained on synthetic or weakly supervised datasets~\cite{fan2018revisiting,bell2014intrinsic,li2021openrooms,roberts2021hypersim}.

More recently, large-scale diffusion models and vision transformers have achieved strong single-view intrinsic predictions. RGB$\leftrightarrow$X~\cite{Zeng2024RGBXID}, Marigold~\cite{ke2023repurposing,Ke2025MarigoldAA}, and PRISM~\cite{Dirik2025PRISMAU} repurpose pretrained generative priors to infer PBR parameters without task-specific retraining. Other approaches explore factorized representations to disentangle diffuse and specular components~\cite{Careaga2024ColorfulDI}, or incorporate multi-modal reasoning to refine estimates~\cite{Dirik2025ReasonXMI}. While these methods produce high-quality per-view predictions, they operate independently on each image and remain sensitive to view-dependent illumination cues, often yielding individually plausible yet mutually inconsistent material estimates across views of the same surface.
\vspace{-.5em}

\subsection{Multi-View and Video Intrinsic Decomposition}
Several lines of work aim to produce consistent intrinsic estimates across multiple views. Inverse rendering pipelines jointly optimize geometry and materials through neural radiance fields or Gaussian splatting under photometric constraints~\cite{Zhang2021NeRFactor,Liang2023GSIR3G,Ye2024GeoSplattingTG}, but require dense captures and costly per-scene optimization. Intrinsic Image Fusion~\cite{Kocsis2025IntrinsicIF} aggregates stochastic single-view predictions into a consistent 3D PBR representation via iterative optimization and path-traced rendering, similarly relying on accurate geometry and dense inputs.

Feed-forward multi-view networks take a different approach. IDT~\cite{Du2025IDTAP} trains a transformer on multi-view inputs to jointly predict reflectance and shading, Event-ID~\cite{Chen2024EventIDID} targets object-centric capture using event cameras, and IDArb~\cite{Li2024IDArbID} trains a diffusion model with cross-view attention on large-scale synthetic object data. While effective for isolated objects under controlled settings, these methods require multi-view training data and specialized architectures, limiting their applicability to scene-level, in-the-wild image collections.

Video-based methods such as Diffusion Renderer~\cite{Liang2025DiffusionRendererNI}, Ouroboros~\cite{Sun2025OuroborosSD} and UniRelight~\cite{He2025UniRelightLJ} exploit temporal coherence to regularize predictions across frames, achieving strong inter-frame consistency. However, they assume dense, ordered sequences and are not directly applicable to sparse, unordered multi-view inputs. In contrast, Geo-ID operates on unordered image collections and repurposes frozen single-view predictors without retraining.

\vspace{-.5em}
\subsection{Diffusion Guidance and Test-Time Optimization}
Guidance mechanisms steer diffusion models at inference time without modifying their weights. Classifier guidance biases sampling using auxiliary objectives~\cite{Dhariwal2021DiffusionMB}, ControlNet~\cite{Zhang2023AddingCC} introduces spatial conditioning during training, and blended diffusion techniques constrain subsets of the latent space to satisfy partial observations~\cite{Avrahami2021BlendedDF}. Test-time optimization has also been explored for inverse problems and image reconstruction~\cite{Tumanyan2022PlugandPlayDF,Kawar2022DenoisingDR}.
Recent work incorporates geometric cues into generative pipelines through camera-aware conditioning or rendering-based supervision~\cite{Shi2023MVDreamMD,Gao2024CAT3DCA,Guo2025MultiviewRV}, but these approaches modify model architectures or require retraining. Geo-ID instead operates purely at inference time, injecting sparse geometric constraints into the denoising process while keeping all model parameters frozen.

\section{Method} \label{sec:method}
Our goal is to produce plausible and cross-view consistent intrinsic decompositions from a sparse, unordered set of images using a pretrained single-view diffusion model, without modifying its parameters. To this end, we introduce an inference-time framework that couples otherwise independent diffusion processes through geometry-guided consensus constraints derived from sparse multi-view correspondences. Our framework, which we denote as Geo-ID, consists of three stages: (i)~geometry-guided correspondence estimation, (ii)~voxel-based consensus initialization, and (iii)~consensus-guided diffusion sampling. An overview of the pipeline is shown in \Cref{fig:pipeline}.

\vspace{-1.5em}
\subsection{Problem Setup}

Let $\{I_i\}_{i=1}^{V}$ denote $V$ images of a static scene. Given a pretrained diffusion model defining a conditional prior
$p_\theta(Y_i \mid I_i)$ over intrinsic maps $Y_i$ (albedo, roughness, and metallicity) for each view independently, our objective is to estimate
$\{\hat{Y}_i\}_{i=1}^{V}$ that remain faithful to the per-view prior while being mutually consistent at corresponding 3D scene locations. All consistency constraints are enforced purely at inference time.

Our goal can be viewed as approximating a joint posterior over intrinsic maps conditioned on all views, while retaining the per-view diffusion prior. Rather than learning a new multi-view model, we approximate this joint objective by enforcing agreement only at geometrically supported 3D correspondences. This yields a sparse, geometry-induced coupling between otherwise independent per-view predictions.

\vspace{-1em}
\paragraph{Base Intrinsic Predictors.}
Our framework is agnostic to the choice of diffusion-based intrinsic predictor and can be applied to any such model. We demonstrate it using two complementary pretrained predictors: (i)~RGB$\leftrightarrow$X~\cite{Zeng2024RGBXID}, which estimates each intrinsic modality via separate text-conditioned diffusion passes, and (ii)~Marigold IID Appearance~\cite{Ke2025MarigoldAA}, which jointly predicts all intrinsic modalities in a single forward pass. 

\vspace{-1em}
\subsection{Geometry-Guided Correspondence Estimation}
\label{subsec:geometry}
\vspace{-0.5em}
We obtain approximate multi-view correspondences and camera geometry using VGGT~\cite{Wang2025VGGTVG}, a feed-forward geometry transformer. For each image $I_i$ of an unordered set of images, VGGT predicts a world-frame point cloud $P_i \in \mathbb{R}^{H \times W \times 3}$, a depth map $D_i \in \mathbb{R}^{H \times W}$, camera extrinsics $E_i$, intrinsics $K_i$, and per-pixel confidence maps $\sigma^P_i$ and $\sigma^D_i$. All 3D quantities are expressed in the coordinate frame of the first camera. We retain points with confidence $\sigma^P_i \geq \tau_c$ (default $\tau_c = 0.35$), and record the originating view $i$ and pixel location for each retained point. When more images are available than the specified number of views, we randomly subsample $V$ images.

\vspace{-1em}
\subsection{Voxel-Based Consensus Initialisation}
\label{subsec:consensus}

After establishing geometric correspondences, we aggregate per-view intrinsic predictions at corresponding 3D locations into robust consensus estimates. Rather than relying on VGGT’s 2D tracking head, we operate directly on the predicted world-frame point clouds, which avoids error accumulation from track propagation, instability arising from visibility scores associated with tracklets, and naturally supports unordered inputs.

We pool high-confidence 3D points from all views into a single point cloud and partition it into axis-aligned voxels of side length $\delta$. When unspecified, we set $\delta = \alpha \cdot \tilde{d}$, where $\tilde{d}$ denotes the median nearest-neighbour distance in the point cloud and $\alpha=2.5$. We discard voxels containing points from fewer than $n_\text{min}$ distinct views (default $n_\text{min}=2$). We then run the base diffusion model independently on each image $I_i$ to obtain initial intrinsic predictions $\hat{Y}^{(0)}_i$. For each voxel $v$, we collect intrinsic values by sampling $\hat{Y}^{(0)}_i$ at the pixel locations corresponding to points that fall inside the voxel.

Let $\mathcal{V}(v)$ denote the set of observations contributing to voxel $v$, and let $u^j_v$ be the image-space location of observation $j$ in its originating view. From the initial per-view predictions $\hat{Y}^{(0)}_i$, we compute a robust consensus estimate and dispersion measure for each voxel:
\vspace{-0.5em}
\begin{equation}
\begin{aligned}
    s_v &= \operatorname{weighted\text{-}median}_{j \in \mathcal{V}(v)}
    \hat{Y}^{(0)}_j(u^j_v), \\
    \hat{\sigma}_v &= 1.4826 \cdot
    \operatorname{median}_{j \in \mathcal{V}(v)}
    \bigl\lvert \hat{Y}^{(0)}_j(u^j_v) - s_v \bigr\rvert .
\end{aligned}
\label{eq:voxel_consensus}
\end{equation}
 
where the weighted median uses observation weights
$w^j_v = \sigma^P_i(u^j_v)\,\log\!\bigl(1 + \lvert \mathcal{V}(v) \rvert\bigr)$ combining the per-point geometric confidence $\sigma^P_i(u^j_v)$ with a term that grows with the number of distinct views $\lvert \mathcal{V}(v) \rvert$ contributing to the voxel. We use the weighted median to ensure robustness to misaligned correspondences or stochastic prediction outliers. The dispersion estimate $\hat{\sigma}_v$ provides a robust measure of cross-view disagreement. First, observations whose deviation from the consensus exceeds a multiple of $\hat{\sigma}_v$ are treated as outliers and excluded from guidance. Second, the inverse variance $\hat{\sigma}_v^{-1}$ modulates the strength of the guidance signal, assigning lower weight to voxels with high intrinsic ambiguity.

To construct per-view guidance targets, we project voxel centers into each camera using the predicted intrinsics and extrinsics, and verify visibility using the predicted depth maps. A voxel is considered visible in view $i$ if its projected pixel location lies within the image bounds and its projected depth $z^i_v$ agrees with the predicted depth $D_i$ within a relative tolerance $\epsilon=0.05$:
\vspace{-0.5em}
\begin{equation}
    m^i_v = \mathbbm{1}\!\left[\,
    \text{in-bounds}(u^i_v) \;\wedge\;
    \frac{\lvert z^i_v - D_i(u^i_v) \rvert}{D_i(u^i_v)} < \epsilon
    \,\right].
    \label{eq:visibility}
\end{equation}
This yields, for each view $i$, a sparse set of visible consensus targets
$\{(u^i_v, s_v, w^i_v)\}$, with weight $w^i_v$ combining the voxel view count and inverse dispersion $\hat{\sigma}_v^{-1}$.

\vspace{-1em}
\subsection{Inference-Time Consensus Guidance}
\label{subsec:guidance}

We obtain the final intrinsic predictions by re-running diffusion sampling for each view while softly enforcing agreement with the consensus targets derived in \Cref{subsec:consensus}. Let $x_{t,i}$ denote the diffusion latent corresponding to view $i$ at denoising timestep $t$. At selected timesteps $t \in \mathcal{S}$, we estimate the corresponding clean intrinsic prediction $\hat{Y}_{t,i}$ from the current latent using the Tweedie formula.
Specifically, given the model’s predicted noise (or $v$-prediction), we recover an estimate of the denoised sample, decode it through the VAE, and evaluate a consistency loss over visible consensus targets:
\begin{equation}
    \mathcal{L}_{t,i}
    =
    \sum_{v:\, m^i_v = 1}
    w^i_v \;
    \rho\!\bigl(\hat{Y}_{t,i}(u^i_v) - s_v\bigr),
    \label{eq:consensus_loss}
\end{equation}
where $s_v$ denotes the voxel consensus value, $u^i_v$ its projected pixel location in view $i$, $w^i_v$ the corresponding confidence weight, and $\rho(\cdot)$ is a Huber loss with threshold $\delta_H = 1.0$. We incorporate this loss by applying a single gradient-based update to the latent,
\begin{equation}
    x_{t,i} \leftarrow x_{t,i} - \eta_t \,\nabla_{x_{t,i}} \mathcal{L}_{t,i},
    \label{eq:latent_update}
\end{equation}
followed by the standard denoising step of the base diffusion scheduler. Only the latent variables are updated; all model parameters remain frozen.

Early diffusion steps primarily determine global structure and coarse material layout, while later steps refine high-frequency detail. Applying guidance during early steps can therefore override structural decisions made by the pretrained prior. We apply guidance only during the last 80\% of denoising steps (i.e., at lower noise levels) and only at voxel-supported pixel coordinates. This allows the generative model to establish coarse structure and material layout in the initial high-noise steps before we enforce cross-view agreement. Although the constraints are shared across views, sampling remains independent per view, avoiding joint multi-image optimization.

\subsection{Implementation Details}
We conduct all experiments on a single NVIDIA A40 GPU. For RGB$\leftrightarrow$X, we use 50 DDIM sampling steps; for Marigold IID Appearance, which is LCM-distilled, we use 10 steps. In both cases, consensus guidance is applied only during the last 80\% of the denoising steps. Visibility is verified using predicted depth with tolerance $\epsilon = 0.05$. All intrinsic predictions and losses are computed in linear RGB space. All reported results use a single fixed configuration ($\alpha=2.5$, $\tau_c=0.35$, $n_\text{min}=2$, $\epsilon=0.05$, $\delta_H=1.0$) across every scene and dataset, with no per-scene tuning. We reserve 20\% of voxels as a held-out set for consistency evaluation.
\vspace{-1em}

\section{Experiments}
\label{sec:experiments}

We evaluate \method along three axes: (i) cross-view intrinsic consistency, (ii) per-view decomposition quality, and (iii) downstream applicability to editable 3D scene representations. We first describe the datasets, baselines, and evaluation metrics used in our study. We then quantify improvements in cross-view agreement across varying numbers of input views, while verifying that per-view decomposition accuracy is preserved. Finally, we demonstrate the practical impact of consistent intrinsics through relightable neural scene reconstruction and conduct ablations to analyze the contribution of key design choices.

\begin{figure*}[t]
    \centering
    \includegraphics[width=\textwidth]{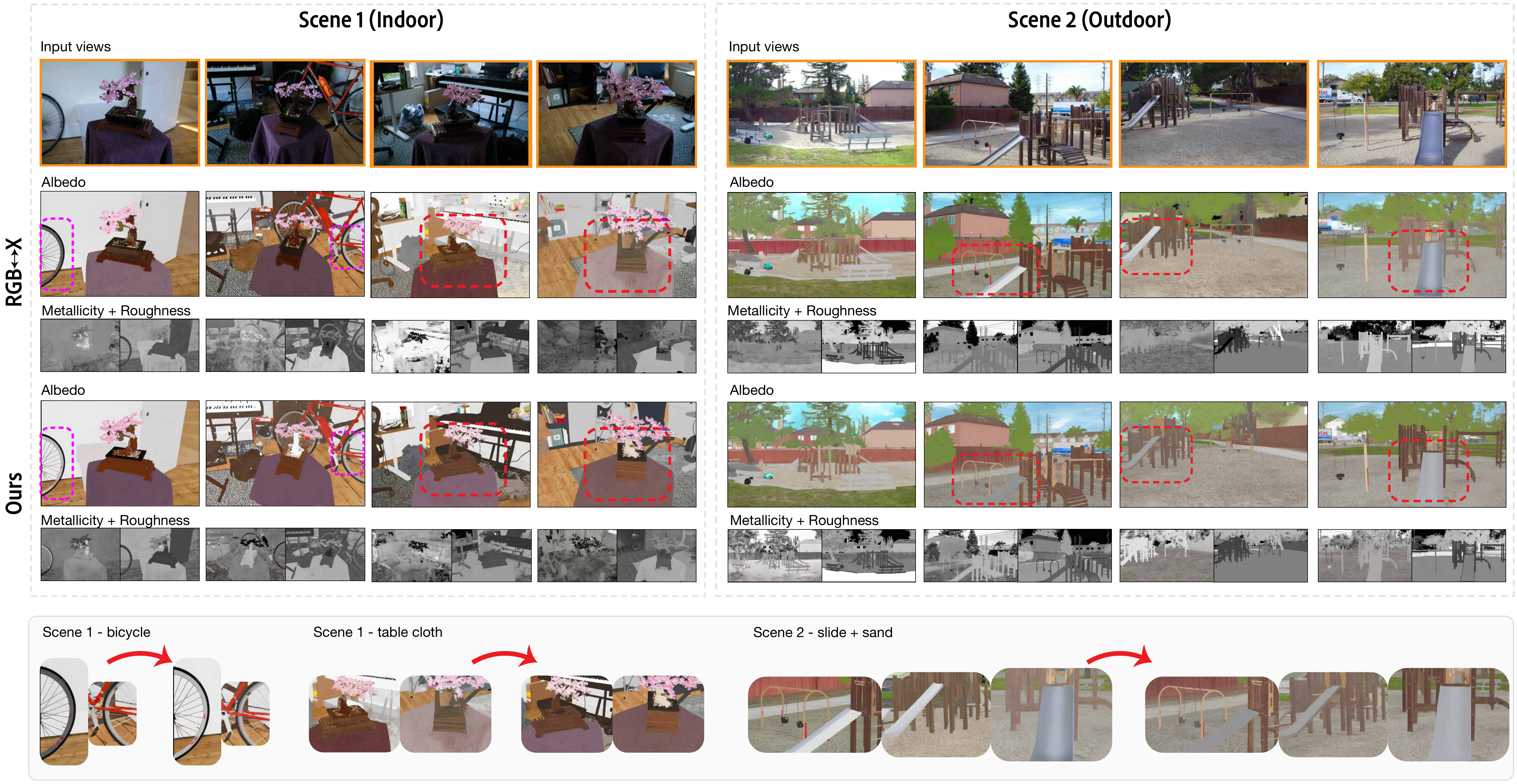}
    \caption{%
        \textbf{Qualitative comparison.} For two scenes (one indoor MipNeRF-360, one outdoor Tanks \& Temples), we show input views alongside albedo, metallicity, and roughness predictions from the base model (RGB$\leftrightarrow$X) applied independently per view and with our \method guidance. Without guidance, the base model produces plausible but inconsistent decompositions: note the color drift across views on the same surfaces (highlighted regions). Bottom row: zoomed-in crops of corresponding surface regions across two views, showing how \method reduces cross-view disagreement while preserving fine detail and decomposition quality.
    }
    \label{fig:qualitative}
    \vspace{-1.5em}
\end{figure*}

\subsection{Experimental Setup}
\label{subsec:exp_setup}

\begin{figure*}[t]
    \centering
    \includegraphics[width=\textwidth]{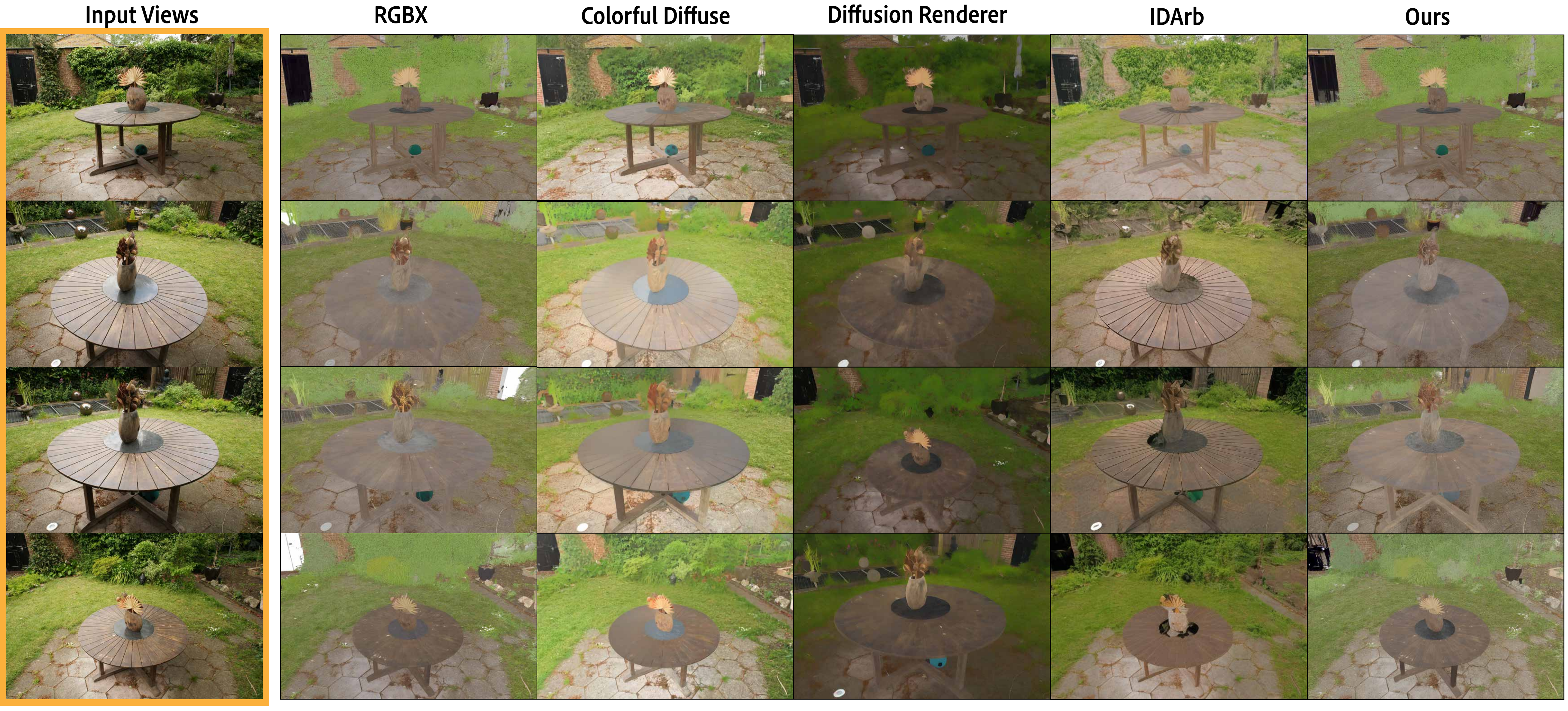}
    \caption{%
    \textbf{Qualitative comparison with single-view, multi-view, and video intrinsic decomposition baselines} on the MipNeRF-360 Garden scene. Each row shows the albedo prediction for a different input view. Single-view methods (RGB$\leftrightarrow$X, IDArb) produce detailed but cross-view inconsistent estimates, while video-based approaches (Diffusion Renderer) generates multi-view consistent results, albeit fails to remove lighting effects. \method (rightmost column, 16-view setting) maintains the detail of single-view predictions while achieving cross-view consistency.
    }
    \label{fig:comparison}
    \vspace{-1.5em}
\end{figure*}

\vspace{-0.5em}
\paragraph{Datasets.}
We evaluate decomposition quality on the synthetic InteriorVerse~\cite{InteriorVerse} and HyperSim~\cite{roberts2021hypersim}. InteriorVerse provides ground-truth albedo, metallicity and roughness maps, whereas HyperSim only provides albedo maps. We use the official test split of HyperSim and follow the same train/test split as RGB$\leftrightarrow$X for InteriorVerse. For cross-view consistency on real-world data, we use scenes from MipNeRF-360~\cite{Barron2021MipNeRF3U} and Tanks\&Temples \cite{Knapitsch2017}. From MipNeRF-360, we evaluate seven scenes (four indoor: \emph{bonsai}, \emph{counter}, \emph{kitchen}, \emph{room}; three outdoor: \emph{bicycle}, \emph{garden}, \emph{stump}), each containing approximately 150 images.
From Tanks\&Temples \cite{Knapitsch2017}, we use the intermediate and advanced scene splits, which provide unordered multi-view captures without ground-truth intrinsics. For each scene and view count $V \in \{4, 8, 16, 32\}$, we sample $k=5$ random view subsets and report averages.  

\vspace{-1em}
\paragraph{Baselines.}
We compare against the unguided single-view predictions of our two base models, RGB$\leftrightarrow$X~\cite{Zeng2024RGBXID} and Marigold IID Appearance~\cite{Ke2025MarigoldAA}, as well as Colorful Diffusion (CID)~\cite{Careaga2024ColorfulDI}, a SOTA single-view model for albedo and irradiance estimation. We additionally compare against the intrinsic decomposition module of the video-based Diffusion Renderer~\cite{Liang2025DiffusionRendererNI}, and evaluate it both in our target unordered, sparse view setting and in its original temporally ordered setting, which serves as our video-based performance upper-bound. All baselines are evaluated using the official model checkpoints and implementations.

\paragraph{Metrics.}
For decomposition quality, we report PSNR, SSIM, and LPIPS for albedo, and RMSE for roughness and metallicity, computed against ground truth on InteriorVerse and HyperSim test sets. For cross-view consistency, we use geometry-aligned correspondences from the held-out 20\% of voxels not used during guidance. For each held-out voxel $v$ observed in views $\mathcal{V}(v)$, we compute the median absolute deviation (MAD) of predicted intrinsic values:
\begin{equation}
    \mathcal{C}_v = \operatorname{MAD}_{i \in \mathcal{V}(v)}\!\bigl(\hat{Y}_i(u^i_v)\bigr).
\end{equation}
We report the average $\mathcal{C}_v$ across held-out voxels, separately for each modality. Lower values indicate better cross-view agreement. For Diffusion Renderer and Colorful Diffuse (CID), which do not use our pipeline, we run VGGT on the same 32-view subsets and evaluate consistency on the same held-out correspondences. 

\vspace{-1em}
\subsection{Cross-View Consistency}
\label{subsec:exp_consistency}
We first evaluate cross-view intrinsic consistency on real-world multi-view datasets and report results on MipNeRF-360 and Tanks\&Temples in \Cref{tab:consistency}. When applied independently to each view, both base models exhibit substantial disagreement across views, particularly for roughness and metallicity, which are more sensitive to view-dependent ambiguities. Geo-ID consistently reduces cross-view disagreement across all modalities and view counts. On MipNeRF-360, consistency improves monotonically with the number of views for both base models. For Marigold Appearance, indoor albedo MAD decreases from 0.091 (unguided) to 0.076 at 32 views, while outdoor metallicity drops from 0.070 to 0.044. RGB$\leftrightarrow$X follows similar trends, with the largest relative gains on roughness and metallicity.  

We evaluate Diffusion Renderer under two settings: \emph{ordered}, where we input the images in their original temporal sequence and read as a video upper bound, and \emph{unordered}, applied to the same randomized subsets as \method. As shown in \Cref{tab:consistency}, removing temporal order degrades the Diffusion Renderer's performance sharply: on MipNeRF-360 outdoor, albedo, roughness, and metallicity MAD rise by roughly $2\times$, $2.4\times$, and $5.4\times$. Being permutation-invariant by construction, our \method beats this matched unordered baseline on albedo across all real-world splits, on every outdoor modality, and on all three InteriorVerse modalities; on the more challenging Tanks\&Temples scenes it leads on albedo and roughness, with metallicity the hardest modality. The ordered upper bound retains an edge only on indoor roughness and metallicity, where dense temporal supervision is strongest. Consistency improvements scale approximately linearly with view count across all modalities and datasets without saturating in the range we evaluate. We note that the gains hold on InteriorVerse with ground-truth correspondences (\Cref{tab:consistency}, rightmost columns), where for Marigold Appearance, albedo MAD drops from 0.085 to 0.070 at 32 views, and hence ruling out mere alignment with VGGT's geometric biases. See the supplementary for per-scene breakdowns and variance across subsets.

\begin{table}[t]
    \caption{%
        Cross-view intrinsic consistency (mean per-correspondence MAD, $\downarrow$). MipNeRF-360 Indoor: \emph{bonsai, counter, kitchen, room}; Outdoor: \emph{bicycle, garden, stump}. Tanks \& Temples and MipNeRF-360 use VGGT correspondences; InteriorVerse uses ground truth. Diffusion Renderer is shown \emph{ordered} (video upper bound) and \emph{unordered} (matched to \method's subsets).
        \method improves consistency across all view counts without retraining.
    }
    \label{tab:consistency}
    \vspace{-0.5em}
    \centering
    \scalebox{0.6}{%
    \begin{tabularx}{1.52\linewidth}{lc | ccc | ccc | ccc | ccc}
        \toprule
        \multirow{2}{*}{\textbf{Method}}
          & \multirow{2}{*}{\textbf{Views}}
          & \multicolumn{3}{c|}{\textbf{Mip-NeRF Ind.}}
          & \multicolumn{3}{c|}{\textbf{Mip-NeRF Out.}}
          & \multicolumn{3}{c|}{\textbf{Tanks \& Temples}}
          & \multicolumn{3}{c}{\textbf{InteriorVerse (GT)}} \\
        \cmidrule(lr){3-5} \cmidrule(lr){6-8} \cmidrule(lr){9-11} \cmidrule(lr){12-14}
          & & Albedo & Rough.\ & Metal.\
          & Albedo & Rough.\ & Metal.\
          & Albedo & Rough.\ & Metal.\
          & Albedo & Rough.\ & Metal.\ \\
        \midrule
        IDArb~\cite{Li2024IDArbID} & 16 & 0.258 & 0.316 & 0.289 & 0.224 & 0.310 & 0.280 & 0.256 & 0.309 & 0.299 & 0.241 & 0.298 & 0.275\\
        CID~\cite{Careaga2024ColorfulDI}
          & 32 & 0.101 & -- & -- & 0.070 & -- & -- & 0.103 & -- & -- & 0.095 & -- & --\\
        Diff.\ Renderer~\cite{Liang2025DiffusionRendererNI} (ordered)
          & 32 & \cg 0.068 & \cg 0.038 & \cg 0.043 & \cg 0.043 & \cg 0.041 & \cg 0.019 & \cg 0.060 & \cg 0.040 & \cg 0.028 & \cg 0.058 & \cg 0.034 & \cg 0.037\\
        Diff.\ Renderer~\cite{Liang2025DiffusionRendererNI} (unordered)
          & 32 & 0.104 & 0.075 & \cI 0.087 & 0.089 & 0.100 & 0.102 & 0.110 & 0.100 & 0.129 & 0.108 & 0.086 & 0.120 \\
        \midrule
        RGB$\leftrightarrow$X~\cite{Zeng2024RGBXID}
          & 32 & \impr0.114 & \impr0.100 & \impr0.206 & \impr0.070 & \impr0.144 & \impr0.192 & \impr0.109 & \impr0.105 & \impr0.219 & \impr0.107 & \impr0.096 & \impr0.198 \\
        Marigold Appr.~\cite{Ke2025MarigoldAA}
          & 32 & \impr0.091 & \impr0.096 & \impr0.111 & \impr0.073 & \impr0.080 & \impr0.070 & \impr0.098 & \impr0.100 & \impr0.118 & \impr0.085 & \impr0.089 & \impr0.103\\
        \midrule
        \multirow{4}{*}{\shortstack[l]{\textbf{RGB$\leftrightarrow$X}\\\textbf{+\ \method}}}
          & 4  & 0.110 & 0.074 & 0.138 & 0.067 & 0.121 & 0.115 & 0.106 & 0.096 & 0.195 & 0.103 & 0.072 & 0.132 \\
          & 8  & 0.107 & \cIII 0.072 & 0.128 & \cIII 0.061 & 0.118 & 0.101 & 0.102 & 0.089 & 0.180 & 0.099 & \cIII 0.068 & 0.121 \\
          & 16 & 0.103 & \cII 0.071 & 0.115 & \cII 0.058 & 0.104 & 0.097 & 0.100 & \cIII 0.082 & 0.174 & 0.095 & \cII 0.065 & 0.109 \\
          & 32 & 0.098 & \cI 0.065 & 0.114 & \cI 0.054 & 0.079 & 0.085 & 0.097 & \cI 0.076 & 0.166 & 0.091 & \cI 0.060 & 0.105 \\
        \midrule
        \multirow{4}{*}{\shortstack[l]{\textbf{Marigold Appr.}\\ \textbf{+\ \method}}}
          & 4  & 0.086 & 0.088 & 0.101 & 0.067 & 0.072 & 0.062 & 0.089 & 0.090 & 0.104 & 0.080 & 0.082 & 0.094\\
          & 8  & \cIII 0.081 & 0.086 & 0.103 & 0.064 & \cIII 0.069 & \cIII 0.059 & \cIII 0.085 & 0.085 & \cIII 0.099 & \cIII 0.075 & 0.078 & \cIII 0.091\\
          & 16 & \cII 0.078 & 0.079 & \cII 0.099 & 0.063 & \cII 0.064 & \cII 0.049 & \cII 0.083 & 0.082 & \cII 0.097 & \cII 0.072 & 0.073 & \cII 0.087\\
          & 32 & \cI 0.076 & 0.082 & \cIII 0.100 & 0.061 & \cI 0.062 & \cI 0.044 & \cI 0.082 & \cII 0.080 & \cI 0.095 & \cI 0.070 & 0.075 & \cI 0.085\\
        \bottomrule
    \end{tabularx}%
    }
    \vspace{-1em}
\end{table}

\vspace{-1.0em}
\subsection{Decomposition Quality}
\label{subsec:exp_quality}
\vspace{-0.5em}

\begin{table}[h]
    \vspace{-1em}
    \caption{%
        Per-view intrinsic decomposition quality on InteriorVerse (left) and HyperSim (right).
        \method achieves cross-view consistency (\Cref{tab:consistency}) without degrading single-view accuracy.
    }
    \label{tab:exp_quality}
    \vspace{-1em}
    \centering
    \scalebox{0.63}{%
    \begin{tabularx}{1.4\linewidth}{lc | *{3}{c} | c | c | *{3}{c}}
        \toprule
        \multirow{3}{*}{\textbf{Method}}
          & \multirow{3}{*}{\textbf{Num. Views}}
          & \multicolumn{5}{c|}{\textbf{InteriorVerse}}
          & \multicolumn{3}{c}{\textbf{HyperSim}} \\
        \cmidrule(lr){3-7} \cmidrule(lr){8-10}
          & & \multicolumn{3}{c|}{\textbf{Albedo}}
            & \textbf{Metallic}
            & \textbf{Roughness}
            & \multicolumn{3}{c}{\textbf{Albedo}} \\
        \cmidrule(lr){3-5} \cmidrule(lr){6-6} \cmidrule(lr){7-7} \cmidrule(lr){8-10}
          & & {\small PSNR~$\uparrow$}
            & {\small SSIM~$\uparrow$}
            & {\small LPIPS~$\downarrow$}
            & {\small RMSE~$\downarrow$}
            & {\small RMSE~$\downarrow$}
            & {\small PSNR~$\uparrow$}
            & {\small SSIM~$\uparrow$}
            & {\small LPIPS~$\downarrow$} \\
        \midrule
        IDArb~\cite{Li2024IDArbID} & 1 & 9.7 & 0.66 & 0.61 & 0.61 & 0.43 & 9.4 & 0.62 & 0.63 \\
        CID \cite{Careaga2024ColorfulDI}                      & 1 & \cIII 17.7 & 0.79 & 0.27 & --   & --   & 17.1 & 0.78 & 0.21 \\
        Kocsis et al. \cite{Kocsis2024IntrinsicID} & 1 & 17.4 & 0.80 & 0.22 & \cIII 0.21 & 0.26 & 12.1 & 0.72 & 0.41 \\
        Diffusion Renderer \cite{Liang2025DiffusionRendererNI}        &  24 & \cI 21.9 & \cI 0.87 & \cI 0.17 & 0.28 & 0.35 & \cI 22.2 & \cIII 0.78 & 0.21 \\
        \midrule
        \rgbx \cite{Zeng2024RGBXID}                             & 1 & 16.4 & 0.78 & \cII 0.19 & 0.44 & 0.38 & 17.4 & \cI 0.81 & \cI 0.18 \\
        Marigold Appr. \cite{Ke2025MarigoldAA}                   & 1 &  \cII 19.5 & \cIII 0.85 & \cII 0.19 & \cII 0.20   & \cIII 0.25   & \cII 18.5 & \cIII 0.78 & \cIII 0.20 \\
        \midrule
        \textbf{RGB$\leftrightarrow$X + \method}         & 4  & 16.2 & 0.77 & \cIII 0.20 & 0.44 & 0.38 & 17.2 & \cII 0.80 & \cII 0.19 \\
                                                     & 8  & 16.3 & 0.78 & \cII 0.19 & 0.42 & 0.36 & 17.3 & \cI 0.81 & \cI 0.18 \\
                                                     & 16 & 16.4 & 0.78 & \cII 0.19 & 0.42 & 0.35 & 17.4 & \cI 0.81 & \cI 0.18 \\
                                                     & 32 & 16.4 & 0.78 & \cII 0.19 & 0.42 & 0.35 & 17.3 & \cI 0.81 & \cI 0.18 \\                                                 
        \midrule
        \textbf{Marigold Appr. + \method}      & 4  & 19.3 & 0.84 & \cIII 0.20 & \cIII 0.21 & 0.26 & 18.3 & 0.77 & 0.21 \\
                                                            & 8  & 19.4 & \cIII 0.85 & \cII 0.19 & \cII 0.20 & \cII 0.24 & \cIII 18.4 & \cIII 0.78 & \cIII 0.20 \\
                                                            & 16 & \cII 19.5 & \cIII 0.85 & \cII 0.19 & \cII 0.20 & \cII 0.24 & \cII 18.5 & \cIII 0.78 &  \cIII 0.20 \\
                                                            & 32 & \cII 19.5 & \cII 0.86 & \cII 0.19 & \cI 0.19 & \cI 0.23 & \cII 18.5 & \cIII 0.78 & \cIII 0.20 \\

        \bottomrule
    \end{tabularx}%
    }
    \vspace{-1.5em}
\end{table}
We next evaluate whether enforcing cross-view consistency degrades per-view intrinsic accuracy. \Cref{tab:exp_quality} shows that \method preserves the decomposition quality of both base models on InteriorVerse and HyperSim. For Marigold Appearance, PSNR, SSIM, and LPIPS remain within ${\pm}0.1$\,dB, ${\pm}0.01$, and ${\pm}0.01$ of the unguided baseline across all view counts; RGB$\leftrightarrow$X exhibits a similar pattern. These results indicate that enforcing cross-view agreement through sparse geometric constraints does not degrade the intrinsic estimates produced by the underlying models.

In some cases (e.g., 16 views), guided predictions slightly improve per-view metrics, indicating that the consensus signal from multiple views can correct isolated single-view errors. Diffusion Renderer~\cite{Liang2025DiffusionRendererNI}, a video-based intrinsic decomposition method, achieves strong per-view accuracy in our evaluation, reflecting the benefits of temporal supervision and joint reasoning across dense frame sequences. However, such approaches assume ordered video input and are trained specifically for multi-frame inference. In contrast, \method is designed as a lightweight inference-time wrapper for existing single-view intrinsic predictors. It introduces sparse geometry-aligned constraints during sampling, enabling cross-view consistency while preserving the original per-view generative prior of the base model.

\subsection{Ablation Studies}
\label{subsec:ablations}

We ablate key design choices on MipNeRF-360 indoor scenes at 16 views using RGB$\leftrightarrow$X as the base model and summarize our results in \Cref{tab:ablation}. We first study the effect of the consensus aggregation strategy. Replacing the weighted median with a weighted mean yields slightly better consistency but at the cost of decomposition quality, as the mean is more sensitive to outlier correspondences that pull predictions away from the base model's prior.

Next, we examine the impact of the guidance schedule. Applying guidance throughout the entire denoising process improves consistency but degrades per-view decomposition quality, suggesting that early-step constraints interfere with structural decisions made by the diffusion prior. Conversely, restricting guidance to only the final 20\% of steps yields weaker consistency gains. Guiding the intermediate-to-late stages provides the best balance between consistency and fidelity.

Finally, we compare our voxel-based 3D correspondence formulation with VGGT’s tracking head, which establishes 2D correspondences by propagating point tracks across views. In this ablation, we replace voxelized 3D aggregation with track-based correspondence and enforce consistency directly at tracked pixel locations. This variant yields worse cross-view consistency, likely because 2D track propagation accumulates drift and relies on reliable visibility ordering, assumptions that are often violated in sparse, unordered multi-view settings. Additional ablation experiments on confidence threshold and voxel size choice, and qualitative examples are provided in the supplementary.

\begin{table*}[b]
\vspace{-1em}
    \caption{%
        Ablation study (16 views, RGB$\leftrightarrow$X).
        Consistency: mean MAD ($\downarrow$) on held-out correspondences (MipNeRF-360 indoor). Quality: decomposition accuracy on the InteriorVerse test set. Aggressive guidance (no outlier rejection, 100\% steps) improves consistency but degrades decomposition quality, while our full model balances both.
    }
    \label{tab:ablation}
    \centering
    \scalebox{0.75}{%
    \begin{tabular}{l ccc | ccc cc}
        \toprule
        & \multicolumn{3}{c|}{\textbf{Consistency (MAD $\downarrow$)}}
        & \multicolumn{5}{c}{\textbf{Decomposition Quality (InteriorVerse)}} \\
        \cmidrule(lr){2-4} \cmidrule(lr){5-9}
        & & & & \multicolumn{3}{c}{Albedo} & Metal.\ & Rough.\ \\
        \cmidrule(lr){5-7} \cmidrule(lr){8-8} \cmidrule(lr){9-9}
        \textbf{Variant}
          & Albedo & Rough.\ & Metal.\
          & PSNR $\uparrow$ & SSIM $\uparrow$ & LPIPS $\downarrow$
          & RMSE $\downarrow$ & RMSE $\downarrow$ \\
        \midrule
        Full model (\method)
          & \textbf{0.103} & \textbf{0.071} & \textbf{0.115}
          & \textbf{16.4} & \textbf{0.78} & \textbf{0.19} & \textbf{0.44} & \textbf{0.38} \\
        \midrule
        Mean consensus (instead of median)
          & 0.098 & 0.066 & 0.099
          & 16.1 & 0.76 & 0.20 & 0.47 & 0.41 \\
        No outlier rejection
          & 0.100 & 0.067 & 0.112
          & 16.2 & 0.75 & 0.20 & 0.45 & 0.41 \\
        Guidance at 100\% of steps
          & 0.096 & 0.067 & 0.106
          & 15.8 & 0.71 & 0.22 & 0.48 & 0.44 \\
        Guidance at last 20\% of steps
          & 0.111 & 0.093 & 0.188
          & 16.4 & 0.78 & 0.19 & 0.44 & 0.38 \\
        2D tracking head (instead of 3D voxels)
          & 0.109 & 0.074 & 0.135
          & 16.2 & 0.75 & 0.21 & 0.47 & 0.42 \\
        \midrule
        Unguided baseline
          & 0.114 & 0.100 & 0.206
          & 16.4 & 0.78 & 0.19 & 0.44 & 0.38 \\
        \bottomrule
    \end{tabular}%
    }
\end{table*}

\begin{table}[t]
    \caption{%
        Runtime analysis on a single NVIDIA A40 GPU (16 views).
        \method adds moderate overhead to the base model, remains competitive with video-based Diffusion Renderer for full PBR prediction, and is orders of magnitude faster than optimisation-based multi-view intrinsic methods. $^\dagger$Requires known geometry and dense captures.
    }
    \label{tab:runtime}
    \vspace{-0.5em}
    \centering
    \scalebox{0.75}{%
    \begin{tabular}{l c c c c}
        \toprule
        & \multicolumn{2}{c}{\textbf{Ours (16 views)}} & & \\
        \cmidrule(lr){2-3}
        \textbf{Component} & Marigold Appr.\ & RGB$\leftrightarrow$X & Diff.\ Renderer & Int.\ Img.\ Fusion$^\dagger$ \\
        \midrule
        Geometry estimation  & 6\,s & 6\,s & -- & known \\
        Unguided init (per view) & 0.9\,s & 3\,s & -- & -- \\
        Consensus computation & 32\,s & 32\,s & -- & -- \\
        Guided pass (per view) & 7.6\,s & 40\,s & -- & -- \\
        Inference (per modality) & -- & -- & 1.77\,min & -- \\
        Per-scene optimisation & -- & -- & -- & $>$1\,hr \\
        \midrule
        \textbf{Total} & \textbf{$\sim$3\,min} & \textbf{$\sim$22\,min} & \textbf{$\sim$5.3\,min} & \textbf{$>$1\,hr} \\
        \bottomrule
    \end{tabular}%
    }
    \vspace{-1em}
\end{table}

\vspace{-1em}
\subsection{Runtime Analysis}
\label{subsec:runtime}

We report per-view and total inference times in \Cref{tab:runtime}. The geometry extraction with VGGT is a one-time cost that amortises over views (3--6s depending on view count). The dominant cost is the guided diffusion pass, which adds ${\sim}30\%$ overhead per view compared to unguided inference due to the VAE decode and gradient computation at guided steps. For full PBR prediction, Diffusion Renderer requires ${\sim}1.77$\,min per modality (${\sim}5.3$\,min total at 16 views). \method with Marigold Appearance, which predicts all three modalities jointly, completes in ${\sim}3$\,min (roughly 40\% faster), while additionally supporting sparse, unordered inputs.

\vspace{-1em}
\subsection{Qualitative Results}
\label{subsec:qualitative}

We present qualitative results on two representative scenes in \Cref{fig:qualitative}: an indoor scene from MipNeRF-360 and an outdoor scene from Tanks\&Temples. We show albedo, roughness, and metallicity predictions from RGB$\leftrightarrow$X applied independently per view alongside predictions refined with \method. For both scenes, we run our method using 16 views. Without guidance, the base model produces plausible per-view decompositions but exhibits clear cross-view inconsistencies, such as color drift on the tablecloth and bicycle frame (Scene 1) and shifting sand and slide tones (Scene 2). We note that \method substantially reduces these discrepancies while preserving fine detail, as the zoomed-in crops in the bottom row confirm.

In \Cref{fig:comparison}, we compare against single-view (RGB$\leftrightarrow$X, IDArb), multi-view (Colorful Diffuse), and video (Diffusion Renderer) intrinsic decomposition baselines on the MipNeRF-360 Garden scene. We observe that single-view methods produce high-quality per-frame predictions but suffer from cross-view inconsistencies, while Diffusion Renderer generates consistent, albeit poorer performance with respect to albedo quality. Our method combines the strengths of both: it maintains the detail of single-view predictions while enforcing cross-view consistency through geometric correspondences. We provide additional examples and failure cases in the supplementary.

\vspace{-1em}
\subsection{Applications}
\label{subsec:exp_applications}
\begin{figure*}[h]
    \centering
    \includegraphics[width=\textwidth]{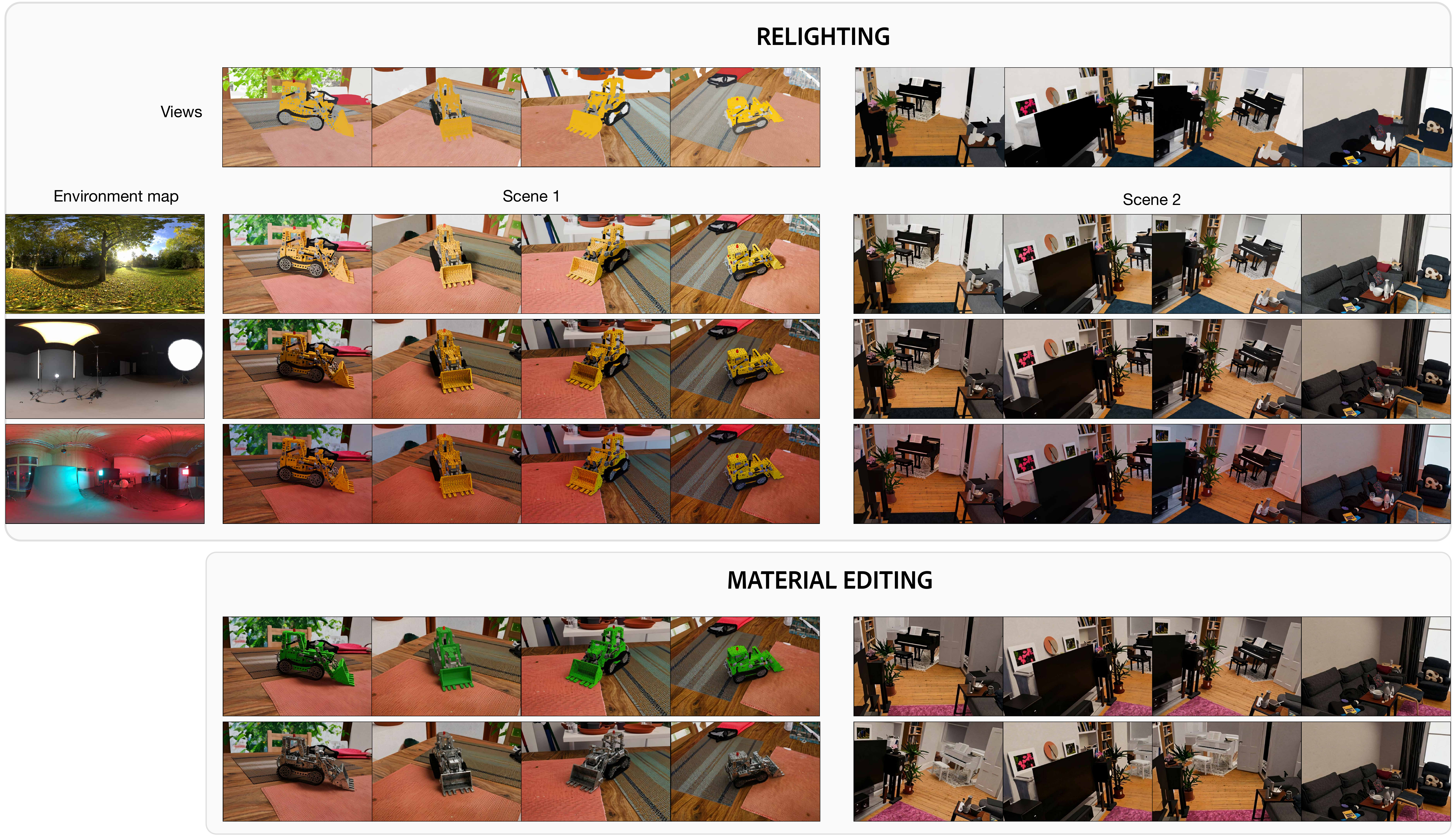}
    \vspace{-1em}
    \caption{%
        \textbf{Downstream applications.} \emph{Top:} We train MeshSplatting~\cite{Held2025MeshSplattingDR} on \method-predicted albedo maps and render the reconstructed surface meshes under novel HDR environment maps. Rows correspond to different lighting conditions; columns show rendered views. \emph{Bottom:} We manually segment target regions on the extracted meshes and modify their albedo to demonstrate material editing. 
    }
    \label{fig:downstream}
    \vspace{-1.5em}
\end{figure*}

We demonstrate the utility of cross-view consistent intrinsics by enabling relightable and editable neural scene representations. To this end, we train a triangular mesh-based neural scene representation using MeshSplatting~\cite{Held2025MeshSplattingDR} on albedo maps predicted by \method for indoor scenes from Mip-NeRF~360. To isolate intrinsic appearance, we replace RGB inputs with predicted albedo maps and disable spherical-harmonic lighting effects by setting $\texttt{sh\_degree}=0$. We follow the original MeshSplatting training setup using monocular depth from Depth Anything~V2~\cite{Yang2024DepthAV} and surface normals from StableNormals~\cite{Ye2024StableNormalRD}, without further modifications.  While a direct quantitative evaluation of relighting quality would require ground-truth intrinsics and lighting that are unavailable for real-world scenes, we quantify the downstream benefit of consistency through albedo novel-view synthesis (\Cref{tab:downstream}) and provide qualitative demonstrations of relighting and material editing in \Cref{fig:downstream}.

\vspace{-0.5em}
\paragraph{Quantitative Downstream Evaluation.}
To quantify how cross-view consistency affects downstream reconstruction, we measure albedo novel-view synthesis. For each method, we train MeshSplatting on a subset of its predicted albedo maps and render at held-out poses, evaluating the rendered views against that method's own predictions at those poses. This deliberately isolates multi-view coherence from absolute correctness: it asks whether a method's per-view albedos form a coherent 3D signal representable by a single editable neural scene, which is the downstream use case our \method method targets, rather than how close they are to ground truth. As reported in \Cref{tab:downstream}, \method improves over its unguided bases by $+1.8$\,dB (RGB$\leftrightarrow$X) and $+1.9$\,dB (Marigold Appearance) in PSNR, with LPIPS reductions of 0.07 and 0.06 respectively. Without guidance, inconsistent per-view albedos provide contradictory supervision for the same 3D surface; \method supplies a more coherent multi-view signal. Both \method variants also outperform ordered Diffusion Renderer, whose residual baked-in lighting (\Cref{fig:comparison}) introduces spurious view-dependence that our view-independent albedo training setup (spherical-harmonic lighting disabled, as described above) cannot absorb.

\begin{table}[t]
    \caption{%
        Downstream albedo novel-view synthesis. We train MeshSplatting~\cite{Held2025MeshSplattingDR} on each method's predicted albedo maps and render at held-out poses, comparing against that method's own held-out predictions. This isolates multi-view coherence from absolute correctness. \method yields the most coherent multi-view albedo signal, outperforming both its unguided bases and ordered Diffusion Renderer.
    }
    \label{tab:downstream}
    \vspace{-0.5em}
    \centering
    \scalebox{0.85}{%
    \begin{tabular}{l ccc}
        \toprule
        \textbf{Training input} & PSNR~$\uparrow$ & SSIM~$\uparrow$ & LPIPS~$\downarrow$ \\
        \midrule
        Diffusion Renderer~\cite{Liang2025DiffusionRendererNI} (ordered) & 20.4 & 0.79 & 0.26 \\
        \midrule
        Unguided RGB$\leftrightarrow$X~\cite{Zeng2024RGBXID} & 19.4 & 0.71 & 0.28 \\
        \textbf{RGB$\leftrightarrow$X + \method} & 21.2 & 0.78 & 0.21 \\
        \midrule
        Unguided Marigold Appr.~\cite{Ke2025MarigoldAA} & 20.1 & 0.74 & 0.25 \\
        \textbf{Marigold Appr. + \method} & \textbf{22.0} & \textbf{0.81} & \textbf{0.19} \\
        \bottomrule
    \end{tabular}%
    }
    \vspace{-1em}
\end{table}
\vspace{-0.5em}
\paragraph{Relighting.}
 We render the reconstructed surface meshes under novel HDR environment maps using physically based rendering in Blender \cite{Hess:2010:BFE:1893021}. As shown in \Cref{fig:downstream}, our \method framework produces consistent albedo across views, the resulting meshes exhibit coherent appearance changes under varying illumination. Inconsistent per-view estimates would instead risk introducing visible seams and color discontinuities, as the mesh would bake conflicting color information into adjacent faces.

\vspace{-0.5em}
\paragraph{Material Editing.}
In addition to relighting, we show that cross-view consistent albedo also enables straightforward scene-level material editing. We manually segment target regions on the extracted meshes and modify their albedo values (e.g., recoloring an object or changing its surface finish). Since \method ensures that the underlying albedo is view-independent, these edits propagate coherently across all rendered viewpoints. Without consistent estimates, such edits would produce view-dependent color artifacts, as each face would encode a different estimate of the same surface.

\vspace{-1em}
\section{Conclusion}
\label{sec:conclusion}
\vspace{-0.5em}
We introduced \method, a training-free framework for enforcing cross-view consistency in intrinsic image decomposition via geometry-guided diffusion at inference time. By coupling independent single-view predictors through sparse geometric correspondences, our approach repurposes pretrained intrinsic models for multi-view settings without retraining or per-scene optimization. Experiments on synthetic benchmarks and real-world scenes show that \method substantially improves cross-view consistency while preserving per-view decomposition quality, and enables downstream applications such as coherent relighting and material editing in neural scene representations.

\vspace{-.5em}
\paragraph{Limitations.} Our approach inherits the failure modes of both its upstream components. First, when the geometry estimator (VGGT) produces severely incorrect correspondences (e.g., in textureless or repetitive regions), the consensus targets become unreliable and can degrade rather than improve predictions. Second, \method cannot correct systematic errors in the base intrinsic predictor — if the underlying model consistently misattributes lighting as albedo, geometric consensus will reinforce rather than resolve the error. Finally, while we demonstrate improvements on both indoor and outdoor scenes, current single-view intrinsic predictors remain biased toward indoor, Lambertian settings, which bounds the quality achievable by any test-time method including ours.

\paragraph{Acknowledgments.} Stefanos Zafeiriou and Alara Dirik have been partially funded by Turing AI Fellowship (EP/Z534699/1).

\bibliographystyle{splncs04}
\bibliography{main}

\appendix
\section{Implementation Details}
\label{sec:supp_implementation}

We provide additional technical details to complement Section~3.5 of the main paper. With both base models, we resize all images to $512{\times}512$ before processing. We optimise the consensus loss (Eq. 3 in the main paper) using Adam~\cite{Kingma2014AdamAM} with a single gradient step per denoising timestep and a constant learning rate: $\eta{=}10$ for RGB$\leftrightarrow$X \cite{Zeng2024RGBXID} and $\eta{=}50$ for Marigold IID Appearance v1.1 \cite{Ke2025MarigoldAA}. The higher rate for Marigold reflects its fewer denoising steps (10 vs.\ 50), which require a stronger per-step correction to achieve comparable cumulative guidance.

For RGB$\leftrightarrow$X, which estimates each intrinsic modality (albedo, roughness, metallicity) via separate text-conditioned diffusion passes, we compute independent consensus targets per modality. Marigold IID Appearance v1.1 predicts all three modalities jointly, so a single consensus map covering all channels is used. In both cases, consensus guidance is applied only during the last 80\% of denoising steps, and all model parameters remain frozen.
\vspace{-.5em}

\section{Additional Results}
\label{sec:supp_quant}
\vspace{-.5em}
\subsection{Per-Scene Cross-View Consistency}
\label{subsec:supp_perscene}
We expand the aggregated consistency results from the main paper (Table~1) into per-scene breakdowns in \cref{tab:supp_mipnerf_indoor,tab:supp_mipnerf_outdoor}, reporting mean MAD~($\downarrow$) and standard deviation across $k{=}5$ random view subsets.

\begin{table*}[h]
    \vspace{-1em}
    \caption{%
        Per-scene cross-view consistency (MAD~$\downarrow$) on \textbf{MipNeRF-360 Indoor} scenes.
        Mean $\pm$ std over $k{=}5$ random view subsets.
    }
    \label{tab:supp_mipnerf_indoor}
    \centering
    \scalebox{0.68}{%
    \begin{tabular}{ll | ccc | ccc | ccc | ccc}
        \toprule
        & & \multicolumn{3}{c|}{\textbf{Bonsai}} & \multicolumn{3}{c|}{\textbf{Counter}} & \multicolumn{3}{c|}{\textbf{Kitchen}} & \multicolumn{3}{c}{\textbf{Room}} \\
        \cmidrule(lr){3-5}\cmidrule(lr){6-8}\cmidrule(lr){9-11}\cmidrule(lr){12-14}
        \textbf{Method} & \textbf{V}
          & Alb. & Rou. & Met.
          & Alb. & Rou. & Met.
          & Alb. & Rou. & Met.
          & Alb. & Rou. & Met. \\
        \midrule
        RGB$\leftrightarrow$X & 32
          & .108{\tiny$\pm$.004} & .092{\tiny$\pm$.003} & .198{\tiny$\pm$.008}
          & .121{\tiny$\pm$.005} & .108{\tiny$\pm$.004} & .215{\tiny$\pm$.009}
          & .118{\tiny$\pm$.004} & .104{\tiny$\pm$.003} & .210{\tiny$\pm$.007}
          & .109{\tiny$\pm$.003} & .096{\tiny$\pm$.004} & .201{\tiny$\pm$.006} \\
        \midrule
        \multirow{4}{*}{\shortstack[l]{RGB$\leftrightarrow$X\\+\,\method}}
          & 4
          & .105{\tiny$\pm$.006} & .070{\tiny$\pm$.005} & .132{\tiny$\pm$.012}
          & .117{\tiny$\pm$.007} & .079{\tiny$\pm$.005} & .145{\tiny$\pm$.014}
          & .113{\tiny$\pm$.005} & .076{\tiny$\pm$.004} & .140{\tiny$\pm$.010}
          & .105{\tiny$\pm$.005} & .071{\tiny$\pm$.004} & .135{\tiny$\pm$.011} \\
          & 8
          & .101{\tiny$\pm$.004} & .068{\tiny$\pm$.003} & .122{\tiny$\pm$.009}
          & .114{\tiny$\pm$.005} & .076{\tiny$\pm$.004} & .135{\tiny$\pm$.010}
          & .110{\tiny$\pm$.004} & .074{\tiny$\pm$.003} & .132{\tiny$\pm$.008}
          & .103{\tiny$\pm$.004} & .070{\tiny$\pm$.003} & .123{\tiny$\pm$.008} \\
          & 16
          & .097{\tiny$\pm$.003} & .067{\tiny$\pm$.002} & .110{\tiny$\pm$.007}
          & .110{\tiny$\pm$.004} & .074{\tiny$\pm$.003} & .121{\tiny$\pm$.008}
          & .106{\tiny$\pm$.003} & .073{\tiny$\pm$.002} & .118{\tiny$\pm$.006}
          & .099{\tiny$\pm$.003} & .069{\tiny$\pm$.002} & .111{\tiny$\pm$.006} \\
          & 32
          & .092{\tiny$\pm$.002} & .062{\tiny$\pm$.002} & .108{\tiny$\pm$.005}
          & .105{\tiny$\pm$.003} & .069{\tiny$\pm$.002} & .120{\tiny$\pm$.006}
          & .101{\tiny$\pm$.002} & .066{\tiny$\pm$.002} & .116{\tiny$\pm$.005}
          & .094{\tiny$\pm$.002} & .063{\tiny$\pm$.002} & .112{\tiny$\pm$.005} \\
        \midrule
        Marigold Appr. & 32
          & .085{\tiny$\pm$.003} & .090{\tiny$\pm$.003} & .105{\tiny$\pm$.005}
          & .097{\tiny$\pm$.004} & .101{\tiny$\pm$.004} & .118{\tiny$\pm$.006}
          & .094{\tiny$\pm$.003} & .098{\tiny$\pm$.003} & .114{\tiny$\pm$.005}
          & .088{\tiny$\pm$.003} & .094{\tiny$\pm$.003} & .107{\tiny$\pm$.005} \\
        \midrule
        \multirow{4}{*}{\shortstack[l]{Marigold Appr.\\+\,\method}}
          & 4
          & .081{\tiny$\pm$.004} & .083{\tiny$\pm$.004} & .096{\tiny$\pm$.006}
          & .092{\tiny$\pm$.005} & .094{\tiny$\pm$.005} & .108{\tiny$\pm$.007}
          & .088{\tiny$\pm$.004} & .090{\tiny$\pm$.004} & .104{\tiny$\pm$.006}
          & .083{\tiny$\pm$.004} & .085{\tiny$\pm$.004} & .098{\tiny$\pm$.006} \\
          & 8
          & .076{\tiny$\pm$.003} & .081{\tiny$\pm$.003} & .098{\tiny$\pm$.005}
          & .088{\tiny$\pm$.004} & .092{\tiny$\pm$.004} & .110{\tiny$\pm$.006}
          & .084{\tiny$\pm$.003} & .088{\tiny$\pm$.003} & .106{\tiny$\pm$.005}
          & .078{\tiny$\pm$.003} & .083{\tiny$\pm$.003} & .099{\tiny$\pm$.005} \\
          & 16
          & .073{\tiny$\pm$.002} & .074{\tiny$\pm$.002} & .094{\tiny$\pm$.004}
          & .084{\tiny$\pm$.003} & .085{\tiny$\pm$.003} & .105{\tiny$\pm$.005}
          & .081{\tiny$\pm$.002} & .081{\tiny$\pm$.002} & .101{\tiny$\pm$.004}
          & .074{\tiny$\pm$.002} & .076{\tiny$\pm$.002} & .096{\tiny$\pm$.004} \\
          & 32
          & .071{\tiny$\pm$.002} & .078{\tiny$\pm$.002} & .095{\tiny$\pm$.004}
          & .082{\tiny$\pm$.003} & .088{\tiny$\pm$.003} & .107{\tiny$\pm$.005}
          & .079{\tiny$\pm$.002} & .084{\tiny$\pm$.002} & .102{\tiny$\pm$.004}
          & .072{\tiny$\pm$.002} & .078{\tiny$\pm$.002} & .096{\tiny$\pm$.004} \\
        \bottomrule
    \end{tabular}%
    }
    \vspace{-1em}
\end{table*}

\begin{table*}[h]
    \caption{%
        Per-scene cross-view consistency (MAD~$\downarrow$) on \textbf{MipNeRF-360 Outdoor} scenes.
        Mean $\pm$ std over $k{=}5$ random view subsets.
    }
    \label{tab:supp_mipnerf_outdoor}
    \vspace{-0.5em}
    \centering
    \scalebox{0.72}{%
    \begin{tabular}{ll | ccc | ccc | ccc}
        \toprule
        & & \multicolumn{3}{c|}{\textbf{Bicycle}} & \multicolumn{3}{c|}{\textbf{Garden}} & \multicolumn{3}{c}{\textbf{Stump}} \\
        \cmidrule(lr){3-5}\cmidrule(lr){6-8}\cmidrule(lr){9-11}
        \textbf{Method} & \textbf{V}
          & Alb. & Rou. & Met.
          & Alb. & Rou. & Met.
          & Alb. & Rou. & Met. \\
        \midrule
        RGB$\leftrightarrow$X & 32
          & .065{\tiny$\pm$.003} & .138{\tiny$\pm$.005} & .185{\tiny$\pm$.008}
          & .072{\tiny$\pm$.003} & .150{\tiny$\pm$.005} & .198{\tiny$\pm$.009}
          & .073{\tiny$\pm$.004} & .144{\tiny$\pm$.006} & .193{\tiny$\pm$.009} \\
        \midrule
        \multirow{4}{*}{\shortstack[l]{RGB$\leftrightarrow$X\\+\,\method}}
          & 4
          & .063{\tiny$\pm$.005} & .116{\tiny$\pm$.007} & .110{\tiny$\pm$.011}
          & .069{\tiny$\pm$.005} & .126{\tiny$\pm$.007} & .121{\tiny$\pm$.012}
          & .069{\tiny$\pm$.006} & .121{\tiny$\pm$.008} & .114{\tiny$\pm$.012} \\
          & 8
          & .058{\tiny$\pm$.003} & .113{\tiny$\pm$.005} & .096{\tiny$\pm$.008}
          & .063{\tiny$\pm$.004} & .122{\tiny$\pm$.006} & .107{\tiny$\pm$.009}
          & .062{\tiny$\pm$.004} & .119{\tiny$\pm$.006} & .100{\tiny$\pm$.009} \\
          & 16
          & .055{\tiny$\pm$.002} & .098{\tiny$\pm$.004} & .092{\tiny$\pm$.006}
          & .060{\tiny$\pm$.003} & .110{\tiny$\pm$.004} & .102{\tiny$\pm$.007}
          & .059{\tiny$\pm$.003} & .104{\tiny$\pm$.005} & .097{\tiny$\pm$.007} \\
          & 32
          & .051{\tiny$\pm$.002} & .074{\tiny$\pm$.003} & .080{\tiny$\pm$.005}
          & .056{\tiny$\pm$.002} & .083{\tiny$\pm$.003} & .090{\tiny$\pm$.005}
          & .055{\tiny$\pm$.002} & .080{\tiny$\pm$.003} & .085{\tiny$\pm$.005} \\
        \midrule
        Marigold Appr. & 32
          & .068{\tiny$\pm$.003} & .075{\tiny$\pm$.003} & .065{\tiny$\pm$.004}
          & .076{\tiny$\pm$.003} & .084{\tiny$\pm$.004} & .074{\tiny$\pm$.004}
          & .075{\tiny$\pm$.003} & .081{\tiny$\pm$.004} & .071{\tiny$\pm$.004} \\
        \midrule
        \multirow{4}{*}{\shortstack[l]{Marigold Appr.\\+\,\method}}
          & 4
          & .062{\tiny$\pm$.004} & .067{\tiny$\pm$.004} & .058{\tiny$\pm$.005}
          & .071{\tiny$\pm$.004} & .076{\tiny$\pm$.005} & .066{\tiny$\pm$.005}
          & .068{\tiny$\pm$.005} & .073{\tiny$\pm$.005} & .062{\tiny$\pm$.005} \\
          & 8
          & .060{\tiny$\pm$.003} & .064{\tiny$\pm$.003} & .055{\tiny$\pm$.004}
          & .067{\tiny$\pm$.003} & .073{\tiny$\pm$.004} & .062{\tiny$\pm$.004}
          & .065{\tiny$\pm$.003} & .070{\tiny$\pm$.004} & .060{\tiny$\pm$.004} \\
          & 16
          & .059{\tiny$\pm$.002} & .060{\tiny$\pm$.002} & .045{\tiny$\pm$.003}
          & .065{\tiny$\pm$.002} & .068{\tiny$\pm$.003} & .053{\tiny$\pm$.003}
          & .064{\tiny$\pm$.003} & .064{\tiny$\pm$.003} & .049{\tiny$\pm$.003} \\
          & 32
          & .057{\tiny$\pm$.002} & .058{\tiny$\pm$.002} & .040{\tiny$\pm$.002}
          & .063{\tiny$\pm$.002} & .066{\tiny$\pm$.002} & .048{\tiny$\pm$.003}
          & .063{\tiny$\pm$.002} & .062{\tiny$\pm$.002} & .044{\tiny$\pm$.002} \\
        \bottomrule
    \end{tabular}%
    }
    \vspace{-.5em}
\end{table*}

We observe that indoor scenes (\emph{bonsai}, \emph{room}) benefit more from guidance than outdoor scenes (\emph{stump}), consistent with VGGT producing higher-quality geometry indoors. Moreover, we observe that variance across random view subsets decreases with the view count, and remains moderate in all settings (typically below 15\% of the mean MAD), confirming that the aggregated numbers in the main paper are representative.

\subsection{Sensitivity to Geometry Quality}
\label{sec:supp_geometry}

A central question is whether \method's gains are bottlenecked by the quality of the geometry estimator. We investigate this on the seven Tanks\&Temples training scenes, which provide laser-scanned ground-truth point clouds and official COLMAP reconstructions with known alignment transforms. We run \method at 32 views with RGB$\leftrightarrow$X. To evaluate our voxelized VGGT predicted geometry, we first align it to the COLMAP reconstruction using the Iterative Closest Point (ICP) algorithm, then apply the official COLMAP-to-ground-truth alignment matrix and crop using the provided bounding boxes. We report the F-score at threshold $\tau{=}0.005 \times d$ (where $d$ is the bounding box diagonal) and the outlier ratio (OR): the fraction of VGGT points whose nearest-neighbour distance to the ground truth exceeds~$2\tau$, in \Cref{tab:supp_geometry}.

\begin{table}[t]
    \caption{%
        VGGT geometry quality vs.\ \method consistency improvement on Tanks\&Temples training scenes (32 views, RGB$\leftrightarrow$X).
        F: F-score ($\uparrow$);
        OR: Outlier Ratio ($\downarrow$).
        $\Delta$MAD: relative reduction in albedo MAD from unguided to guided (\%).
    }
    \label{tab:supp_geometry}
    \vspace{-0.5em}
    \centering
    \scalebox{0.78}{%
    \begin{tabular}{l | cc | cc | c}
        \toprule
        \textbf{Scene}
          & F ($\uparrow$) & OR ($\downarrow$)
          & MAD$_{\text{base}}$ & MAD$_{\text{ours}}$
          & $\Delta$MAD (\%) \\
        \midrule
        Barn          & 0.82 & 0.10 & .104 & .091 & $-$12.5 \\
        Caterpillar   & 0.75 & 0.15 & .108 & .096 & $-$11.1 \\
        Church        & 0.54 & 0.32 & .116 & .108 & $-$6.9 \\
        Courthouse    & 0.48 & 0.38 & .120 & .114 & $-$5.0 \\
        Ignatius      & 0.58 & 0.29 & .106 & .094 & $-$11.3 \\
        Meeting Room  & 0.71 & 0.17 & .110 & .099 & $-$10.0 \\
        Truck         & 0.68 & 0.21 & .112 & .103 & $-$8.0 \\
        \bottomrule
    \end{tabular}%
    }
    \vspace{-1em}
\end{table}

We observe a clear trend: scenes with accurate geometry (Barn, Caterpillar, Ignatius) see the largest consistency gains ($>$10\% relative MAD reduction), while scenes with weaker reconstructions (Courthouse, Church) yield more modest improvements. Notably, \method does not degrade consistency in any scene, confirming that outlier rejection and confidence weighting provide adequate safeguards under noisy geometry. The Pearson correlation between F-score and relative MAD improvement is $r{=}0.94$, suggesting that advances in feed-forward geometry estimation will directly translate into stronger consistency gains.

\section{Ablation Studies}
\label{sec:supp_ablations}
\vspace{-.5em}

\subsection{Confidence Threshold $\tau_c$}
\label{subsec:supp_tau}

We vary the VGGT confidence threshold $\tau_c$ used to filter 3D points before voxelisation. \Cref{tab:supp_tau} reports results on MipNeRF-360 indoor scenes at 16 views (RGB$\leftrightarrow$X).

\begin{table}[t]
    \caption{%
        Ablation on confidence threshold $\tau_c$ (16 views, RGB$\leftrightarrow$X, MipNeRF-360 Indoor). Default $\tau_c{=}0.35$ provides the best consistency--quality tradeoff.
    }
    \label{tab:supp_tau}
    \vspace{-0.5em}
    \centering
    \scalebox{0.78}{%
    \begin{tabular}{c | ccc | ccc}
        \toprule
        & \multicolumn{3}{c|}{\textbf{Consistency (MAD $\downarrow$)}}
        & \multicolumn{3}{c}{\textbf{Quality (InteriorVerse)}} \\
        \cmidrule(lr){2-4}\cmidrule(lr){5-7}
        $\tau_c$ & Albedo & Rough. & Metal. & PSNR $\uparrow$ & SSIM $\uparrow$ & LPIPS $\downarrow$ \\
        \midrule
        0.10 & 0.115 & 0.103 & 0.200 & 15.9 & 0.74 & 0.21 \\
        0.20 & 0.107 & 0.090 & 0.146 & 16.1 & 0.76 & 0.20 \\
        \textbf{0.35} & \textbf{0.103} & \textbf{0.071} & \textbf{0.115} & \textbf{16.4} & \textbf{0.78} & \textbf{0.19} \\
        0.50 & 0.106 & 0.075 & 0.123 & 16.4 & 0.78 & 0.19 \\
        0.70 & 0.110 & 0.084 & 0.152 & 16.4 & 0.78 & 0.19 \\
        \bottomrule
    \end{tabular}%
    }
\end{table}

Low thresholds ($\tau_c \leq 0.20$) admit noisy correspondences that degrade both consistency and decomposition quality, as unreliable consensus targets pull predictions away from the base model's prior. High thresholds ($\tau_c \geq 0.50$) produce overly sparse consensus maps with insufficient constraints to reduce cross-view disagreement. The default $\tau_c{=}0.35$ balances correspondence density and reliability.

\vspace{-.5em}
\subsection{Voxel Size $\delta$}
\label{subsec:supp_voxel}
We control voxel size via the multiplier $\alpha$ in $\delta = \alpha \cdot \tilde{d}$, where $\tilde{d}$ is the median nearest-neighbour distance. Next, we ablate the effect of voxel size on consistency and decomposition performance for $\alpha \in \{1.0, 1.5, 2.5, 4.0, 6.0\}$ and present the results in \Cref{tab:supp_voxel}.

\begin{table}[h]
\vspace{-1.0em}
    \caption{%
        Ablation on voxel size multiplier $\alpha$ (16 views, RGB$\leftrightarrow$X, MipNeRF-360 Indoor).
        Default $\alpha{=}2.5$ balances spatial resolution with observation count per voxel.
    }
    \label{tab:supp_voxel}
    \vspace{-0.5em}
    \centering
    \scalebox{0.75}{%
    \begin{tabular}{c | ccc | ccc | c}
        \toprule
        & \multicolumn{3}{c|}{\textbf{Consistency (MAD $\downarrow$)}}
        & \multicolumn{3}{c|}{\textbf{Quality (InteriorVerse)}}
        & \textbf{Avg.\ obs.} \\
        \cmidrule(lr){2-4}\cmidrule(lr){5-7}
        $\alpha$ & Albedo & Rough. & Metal. & PSNR $\uparrow$ & SSIM $\uparrow$ & LPIPS $\downarrow$ & per voxel \\
        \midrule
        1.0 & 0.106 & 0.078 & 0.132 & 16.3 & 0.77 & 0.19 & 2.4 \\
        1.5 & 0.104 & 0.074 & 0.122 & 16.3 & 0.78 & 0.19 & 3.8 \\
        \textbf{2.5} & \textbf{0.103} & \textbf{0.071} & \textbf{0.115} & \textbf{16.4} & \textbf{0.78} & \textbf{0.19} & \textbf{6.2} \\
        4.0 & 0.101 & 0.069 & 0.112 & 16.2 & 0.77 & 0.20 & 10.5 \\
        6.0 & 0.100 & 0.068 & 0.110 & 16.0 & 0.75 & 0.20 & 18.1 \\
        \bottomrule
    \end{tabular}%
    }
    \vspace{-1.5em}
\end{table}

As shown in \Cref{tab:supp_voxel}, small voxels ($\alpha{=}1.0$) contain too few observations -- often from a single view -- to form a meaningful consensus, whereas large voxels ($\alpha \geq 4.0$) mix observations from spatially distant surfaces, introducing blurring that degrades decomposition quality. We note that the quality degradation at $\alpha{=}6.0$ remains modest despite a threefold increase in observations per voxel (18.1 vs.\ 6.2), which we attribute to the Huber loss and outlier rejection dampening the effect of mixed-surface voxels. The default $\alpha{=}2.5$ yields 6.2 observations per voxel on average, sufficient for robust median estimation while preserving spatial precision and preventing loss of detail.

\vspace{-1em}
\section{Qualitative Results}
\label{sec:supp_qualitative}
\vspace{-1em}

\begin{figure*}[h]
    \vspace{-1.5em}
    \centering
    \includegraphics[width=\textwidth]{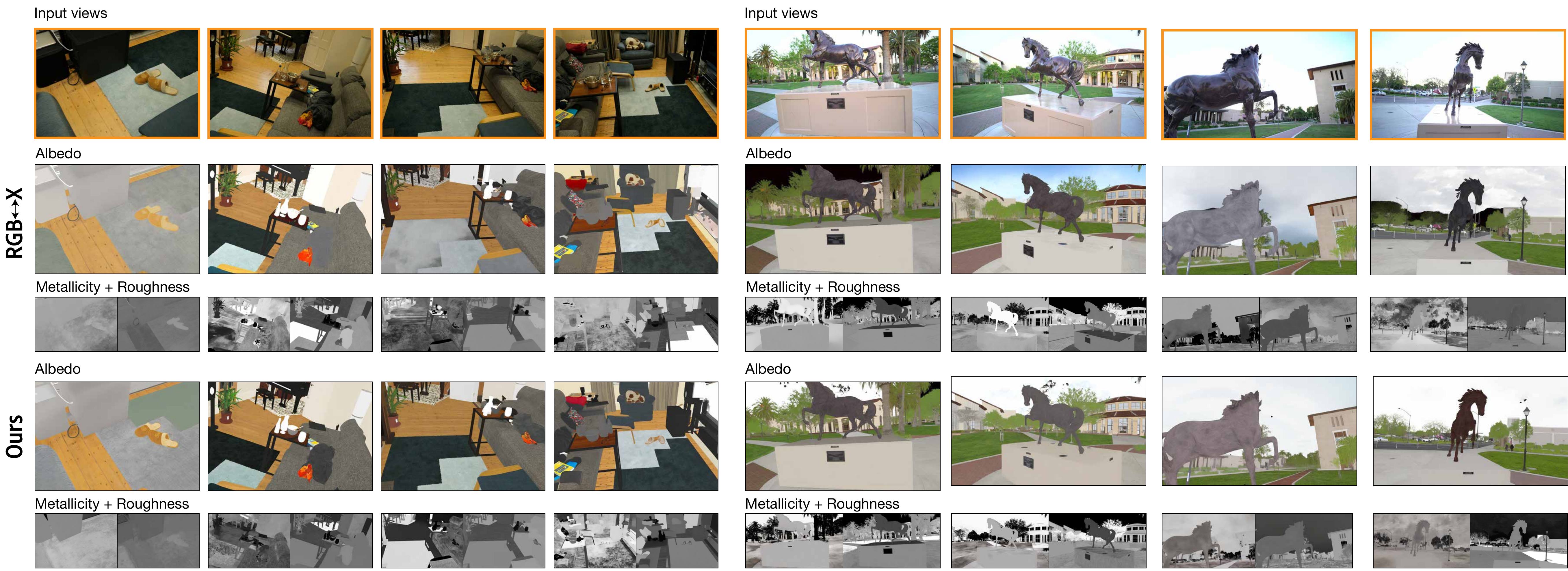}
    \caption{%
        Additional qualitative results on an indoor MipNeRF-360 (left) and an outdoor Tanks\&Temples (right) scene. For each scene, we show unguided (top) and \method-guided (bottom) RGB$\leftrightarrow$X predictions across multiple views.
    }
    \label{fig:supp_qual1}
\end{figure*}

\begin{figure*}[h]
    \vspace{-1em}
    \centering
    \includegraphics[width=\textwidth]{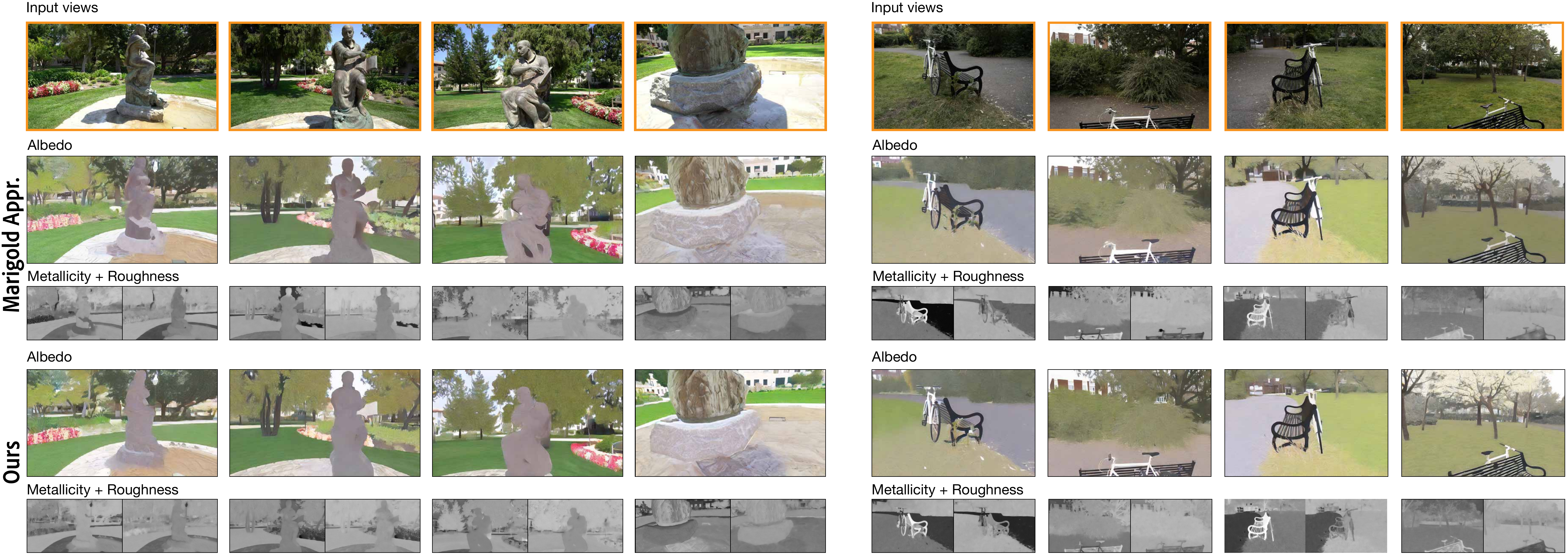}
    \caption{%
        Additional qualitative results on Tanks\&Temples (left) and MipNeRF-360 (right) with Marigold Appearance and our \method-guided version. \method eliminates view-dependent lighting artifacts (\eg, on the bench and sculpture surfaces) and substantially stabilises metallicity and roughness predictions.
    }
    \label{fig:supp_qual2}
    \vspace{-3em}
\end{figure*}

We provide additional qualitative comparisons in \cref{fig:supp_qual1,fig:supp_qual2} with RGB$\leftrightarrow$X and Marigold IID Appearance v1.1 as base models respectively, and present results for MipNeRF-360 and Tanks\&Temples scenes not shown in the main paper. All \method samples shown in the figures are generated in a 16-view setting. We note that the improvements are consistent with the quantitative and qualitative trends in our main paper: \method reduces visible color drift and material inconsistencies across views while preserving fine detail.

\vspace{-2em}

\section{Consensus Target Visualisation}
\label{sec:supp_consensus_viz}

\begin{figure}[h]
    \vspace{-1em}
    \centering
    \includegraphics[width=\textwidth]{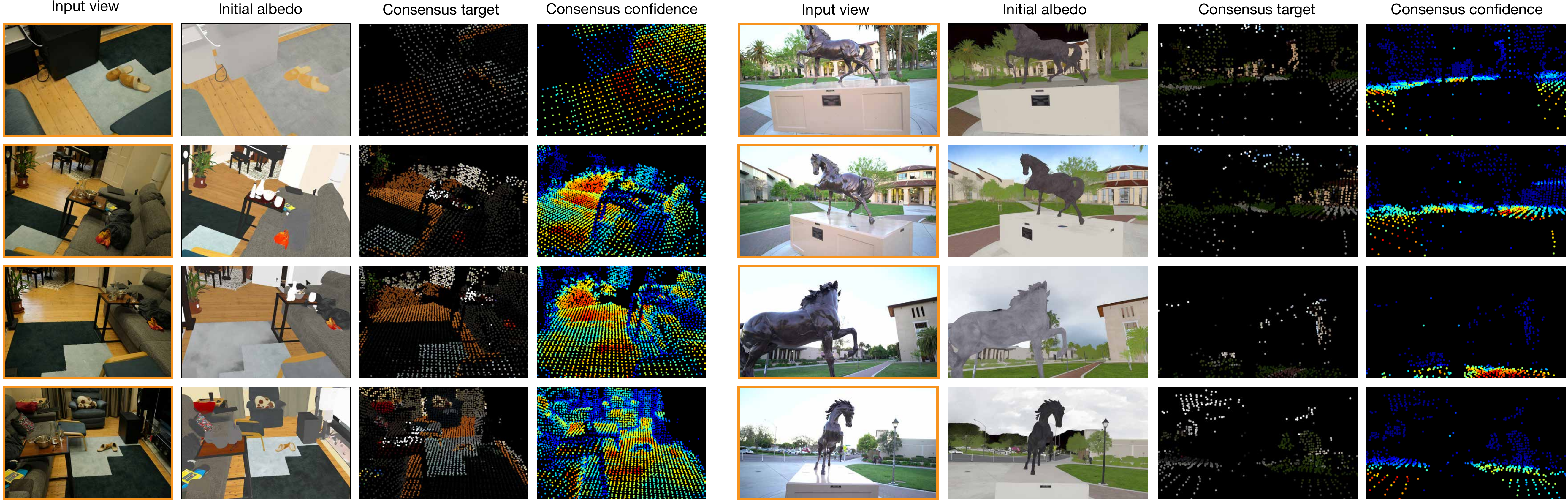}
    \caption{%
        Projected consensus targets for representative scenes.
        \textbf{Left:} input view. \textbf{Middle:} consensus albedo values at projected voxel locations. \textbf{Right:} inverse dispersion weights $\hat{\sigma}_v^{-1}$ (brighter = higher confidence). Indoor scenes produce denser, higher-confidence targets than outdoor scenes, consistent with the stronger consistency improvements observed in \cref{tab:supp_mipnerf_indoor,tab:supp_mipnerf_outdoor}.
    }
    \label{fig:supp_consensus}
    \vspace{-1em}
\end{figure}

To build intuition about the density and spatial distribution of the consensus guidance signal, we visualize the projected albedo consensus targets for representative scenes in \cref{fig:supp_consensus}. As expected, indoor scenes produce denser, higher-confidence targets than outdoor scenes, where points are more spread out and some objects receive only sparse coverage. Interestingly, even regions with few consensus targets -- such as the horse sculpture in the outdoor scene -- still show improved consistency in practice (\cref{fig:supp_qual1}), likely because the consensus loss gradient in latent space propagates spatially beyond the exact target locations through the U-Net's convolutional and attention layers.

\begin{figure*}[t]
    \centering
    \includegraphics[width=\textwidth]{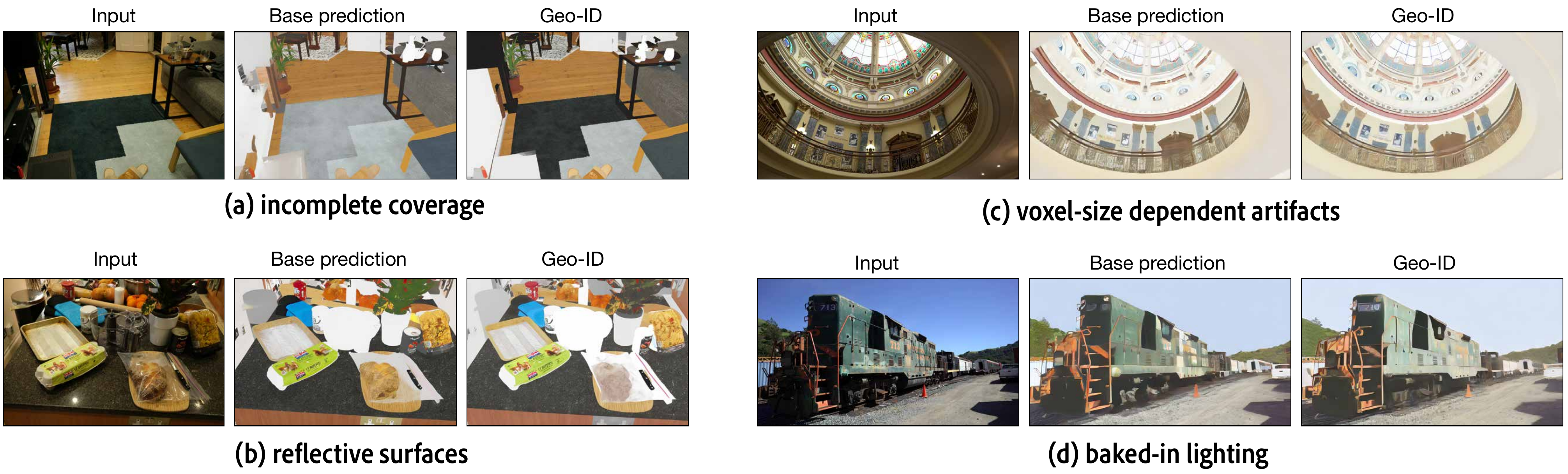}
    \caption{%
        Failure cases of \method. Each triplet shows the input image (left), unguided base prediction (middle), and \method-refined prediction (right). Cases~(a,\,b) use RGB$\leftrightarrow$X; cases~(c,\,d) use Marigold Appearance.
        \textbf{(a)}~Incomplete consensus coverage: the dark metal and glass TV stand lacks high-confidence VGGT correspondences, leaving the prediction unchanged.
        \textbf{(b)}~Reflective surfaces: the base model fails on glossy materials with significantly varying initial view predictions, and consensus propagates these errors across views.
        \textbf{(c)}~Voxel-size artifacts: fine details (railings, ornamental trim) fall below the voxel resolution, causing localized smudging.
        \textbf{(d)}~Baked-in lighting: shadows on dark metallic surfaces are misattributed to albedo by the base model, and consensus reinforces this across views.
    }
    \label{fig:supp_failures}
    \vspace{-1em}
\end{figure*}

\section{Failure Cases}
\label{sec:supp_failures}

Finally, we identify four characteristic failure modes with our method, illustrated in \cref{fig:supp_failures}. The first is incomplete consensus coverage: when VGGT assigns low confidence to a region (\eg, the dark metal and glass TV stand in \cref{fig:supp_failures}a), too few points survive filtering to form consensus targets and the prediction is left unchanged. We argue that this constitutes a graceful degradation: consistency is not improved, but decomposition quality is preserved.

A related but more harmful case arises on reflective and specular surfaces (\cref{fig:supp_failures}b), where both the geometry estimator and the base intrinsic predictor produce unreliable outputs, with highly variable initial predictions across views. Here, \method's consensus can propagate erroneous predictions across views rather than correcting them. A third failure mode is voxel-size dependent: fine geometric details smaller than the voxel resolution~$\delta$ may be merged with adjacent surfaces, causing the consensus to average disparate intrinsic values and produce localized smudging (\cref{fig:supp_failures}c). Reducing~$\alpha$ mitigates this at the cost of weaker consensus overall (see \cref{tab:supp_voxel}).

Finally, when the base predictor systematically misattributes strong shadows to albedo, as we show for Marigold Appearance v1.1 on dark metallic surfaces (\cref{fig:supp_failures}d), these errors are consistent across nearby viewpoints, and geometric consensus reinforces rather than corrects them. This limitation is fundamental to any test-time consistency method that preserves the base model's prior without modifying its weights.

\nopagebreak
\end{document}